\documentclass[11pt]{article}
\pdfoutput=1 

\newif\ifDraftMode
\newif\ifAnonymousSubmissionMode
\newif\ifArxivMode

\DraftModefalse
\AnonymousSubmissionModefalse
\ArxivModetrue

\ifAnonymousSubmissionMode
    \usepackage[review]{acl}
\else
    \ifArxivMode
        \usepackage[preprint]{acl}
    \else
        \usepackage[review]{acl}
    \fi
\fi

\usepackage{times}
\usepackage{latexsym}
\usepackage[T1]{fontenc}
\usepackage[utf8]{inputenc}
\usepackage{microtype}
\usepackage{inconsolata}

\usepackage{array}
\usepackage{amsfonts}
\usepackage{amsmath}
\usepackage{amssymb}
\usepackage{amsthm}
\usepackage{bbm}
\usepackage{bm}
\usepackage{booktabs}
\usepackage[thicklines]{cancel}
\usepackage{changepage}
\usepackage{emoji}
\usepackage[inline]{enumitem}
\usepackage{float}
\usepackage{forest}
\usepackage{mathtools}
\usepackage{pgfplots}
\pgfplotsset{compat=1.18}
\usepackage{siunitx}
\usepackage{subcaption}
\usepackage{tikz}
\usetikzlibrary{arrows.meta}
\usetikzlibrary{positioning}
\usepackage[normalem]{ulem}
\usepackage{url}
\usepackage[dvipsnames,table]{xcolor}

\usepackage[capitalize,noabbrev]{cleveref}

\definecolor{ETHBlue}{RGB}{33,92,175}	
\definecolor{ETHGreen}{RGB}{98,115,19}		
\definecolor{ETHPurpleDark}{RGB}{140,10,89}	
\definecolor{ETHPurple}{RGB}{163,7,116}	
\definecolor{ETHGray}{RGB}{111,111,111}	
\definecolor{ETHRed}{RGB}{183,53,45}	
\definecolor{ETHPetrol}{RGB}{0,120,148}	
\definecolor{ETHBronze}{RGB}{142,103,19}	

\colorlet{ETHdarkblue}{ETHBlue!80!black}
\colorlet{ETHdarkgreen}{ETHGreen!80!black}
\colorlet{ETHpink}{ETHPurple}
\colorlet{ETHgray}{ETHGray}
\colorlet{ETHred}{ETHRed}
\colorlet{ETHgreenblue}{ETHPetrol}
\colorlet{ETHbrown}{ETHBronze}

\definecolor{TextBlack}{RGB}{51,51,51}
\definecolor{BackgroundWhite}{RGB}{255,255,255}

\definecolor{AccentBlue}{RGB}{0,122,204}
\definecolor{LightBlue}{RGB}{173,216,230}
\definecolor{DarkBlue}{RGB}{0,51,102}

\definecolor{AccentGreen}{RGB}{70,160,73}
\definecolor{LightGreen}{RGB}{144,238,144}
\definecolor{DarkGreen}{RGB}{0,100,0}

\definecolor{AccentRed}{RGB}{255,0,0}
\definecolor{LightRed}{RGB}{255,99,71}
\definecolor{DarkRed}{RGB}{139,0,0}

\definecolor{AccentOrange}{RGB}{255,165,0}
\definecolor{LightOrange}{RGB}{255,204,153}
\definecolor{DarkOrange}{RGB}{255,140,0}

\definecolor{NeutralLightGray}{RGB}{204,204,204}
\definecolor{NeutralMediumGray}{RGB}{102,102,102}

\definecolor{NoteYellow}{RGB}{255,255,0}

\definecolor{DiversePurple}{RGB}{128,0,128}
\definecolor{DiverseTeal}{RGB}{0,128,128}
\definecolor{DiverseOlive}{RGB}{128,128,0}
\definecolor{DiverseCyan}{RGB}{0,128,192}
\definecolor{DiverseMagenta}{RGB}{192,0,128}

\colorlet{MacroColor}{ETHPetrol}
\colorlet{MACROCOLOR}{MacroColor}

\AddToHook{cmd/appendix/before}{\crefalias{section}{appendix}}

\theoremstyle{plain}

\theoremstyle{definition}

\theoremstyle{remark}

\pgfplotsset{
    mybarchart/.style={
        ybar,
        width=7.8cm,
        height=3.5cm,
        ymin=0,
        ymax=1,
        bar width=10pt,
        xtick=data,
        error bars/y dir=both,
        error bars/y explicit,
        title style={
            at={(0.5,1)},
            yshift=-5pt,
            anchor=south,
        },
        xtick pos=bottom,
        ytick pos=left,
        xticklabel style={
            rotate=20,
            anchor=east,
        },
    },
    hierarchical/.style={
        draw=green!60!black,
        fill=green!40,
        error bars/error bar style={color=green!60!black},
    },
    linear/.style={
        draw=red!60!black,
        fill=red!40,
        error bars/error bar style={color=red!60!black},
    }
}

\ifAnonymousSubmissionMode
    \newcommand{\macro}[1]{#1}
\else
    \ifArxivMode
        \newcommand{\macro}[1]{#1}
    \else
        \newcommand{\macro}[1]{\textcolor{ETHPetrol}{#1}}
    \fi
\fi

\newcommand{\newterm}[1]{\textbf{#1}}

\newcommand{\varPart}[1]{\ensuremath{_{\pm#1}}}
\newcommand{\meanAndVar}[2]{#1 & \varPart{#2}}
\newcommand{\meanAndVarBold}[2]{\bfseries #1 & \varPart{#2}}

\newcommand{\linguisticRuleName}[1]{\textbf{\textsc{#1}}}
\newcommand{\specialToken}[1]{\textsc{#1}}
\newcommand{\textDelete}[1]{\textcolor{red}{\sout{#1}}}
\newcommand{\textInsert}[1]{\textcolor{Green}{#1}}

\newcommand{\moveMain}{\linguisticRuleName{move-main}}
\newcommand{\moveFirst}{\linguisticRuleName{move-first}}
\newcommand{\qfDecl}{\macro{\specialToken{decl}}}
\newcommand{\qfQuest}{\macro{\specialToken{quest}}}

\newcommand{\agreeMain}{\linguisticRuleName{agree-main}}
\newcommand{\agreeRecent}{\linguisticRuleName{agree-recent}}
\newcommand{\trPast}{\macro{\specialToken{past}}}
\newcommand{\trPresent}{\macro{\specialToken{present}}}

\newcommand{\moveObject}{\linguisticRuleName{move-object}}
\newcommand{\moveSecond}{\linguisticRuleName{move-second}}
\newcommand{\passDecl}{\macro{\specialToken{decl}}}
\newcommand{\passPassiv}{\macro{\specialToken{passiv}}}

\newcommand{\transformerLabel}{Tf}
\newcommand{\rnnLabel}{RNN}
\newcommand{\lstmLabel}{LSTM}
\newcommand{\supLabel}{Sup}
\newcommand{\ndLabel}{Nd}
\newcommand{\readingLabel}{R}

\newcommand{\treeFootnote}{\emoji[emojis]{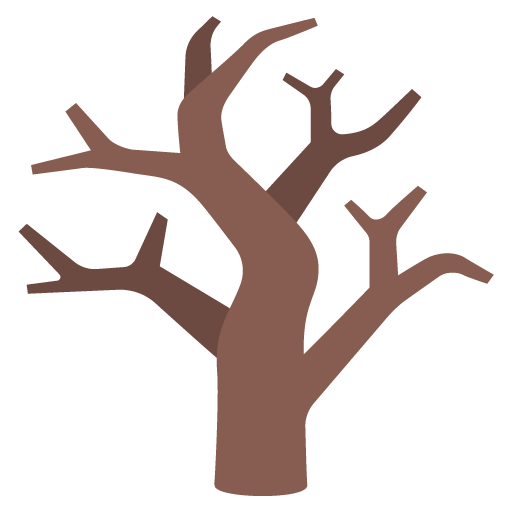}}
\newcommand{\treeFootnoteText}{\treeFootnote{} Uses ground-truth parse trees.}
\newcommand{\pretrainedFootnote}{\emoji[emojis]{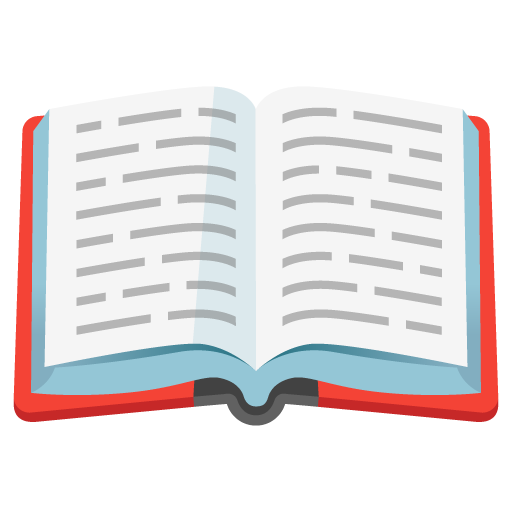}}
\newcommand{\pretrainedFootnoteText}{\pretrainedFootnote{} Large pre-trained model.}
\newcommand{\overtrainedFootnote}{\emoji[emojis]{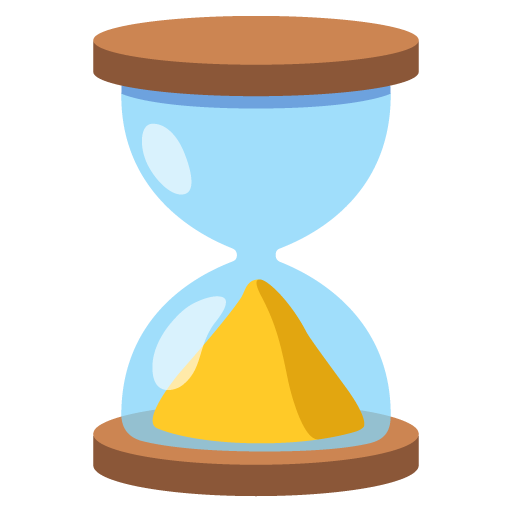}}
\newcommand{\overtrainedFootnoteText}{\overtrainedFootnote{} Overtrained.}

\newcommand{\mccoyFootnote}{\textsuperscript{a}}
\newcommand{\mccoyFootnoteText}{\mccoyFootnote{}Results from \citet{mccoy-etal-2020-syntax}.}
\newcommand{\muellerFootnote}{\textsuperscript{e}}
\newcommand{\muellerFootnoteText}{\muellerFootnote{}Results from \citet{mueller-etal-2022-coloring}.}
\newcommand{\murtyFootnote}{\textsuperscript{b}}
\newcommand{\murtyFootnoteText}{\murtyFootnote{}Approximate results from \citet{murty-etal-2023-grokking}.}
\newcommand{\ahujaFootnote}{\textsuperscript{c}}
\newcommand{\ahujaFootnoteText}{\ahujaFootnote{}Approximate results from \citet{ahuja-etal-2025-learning} using the language modeling objective.}
\newcommand{\qinFootnote}{\textsuperscript{d}}
\newcommand{\qinFootnoteText}{\qinFootnote{}Approximate results from \citet{qin-etal-2025-data}.}

\newcommand{\vecVar}[1]{\bm{#1}}
\newcommand{\matVar}[1]{\bm{#1}}
\newcommand{\stringVar}[1]{\bm{#1}}
\newcommand{\funcName}[1]{\textsc{#1}}
\newcommand{\zeroVector}{\macro{\bm{0}}}

\newcommand{\defeq}{\mathrel{\stackrel{\textnormal{\tiny def}}{=}}}
\newcommand{\realSet}{\macro{\mathbb{R}}}
\newcommand{\positiveRealSet}{\realSet_{+}}

\newcommand{\logistic}[1]{\sigma(#1)}
\newcommand{\elementwiseMultiply}{\mathbin{\macro{\odot}}}

\newcommand{\emptyString}{\macro{\varepsilon}}

\newcommand{\bos}{\macro{\textsc{bos}}}
\newcommand{\eos}{\macro{\textsc{eos}}}

\newcommand{\vectorParam}[2]{\vecVar{#1}_{\mathrm{#2}}}
\newcommand{\vectorParamL}[3]{\vectorParam{#1}{#2}^{(#3)}}
\newcommand{\linearTransform}[2]{\weightParam{#1} #2}
\newcommand{\linearTransformL}[3]{\weightParamL{#1}{#2} #3}
\newcommand{\affine}[2]{\linearTransform{#1}{#2} + \biasParam{#1}}
\newcommand{\affineL}[3]{\linearTransformL{#1}{#2}{#3} + \biasParamL{#1}{#2}}
\newcommand{\weightParamLetter}{\macro{\matVar{W}}}
\newcommand{\weightParam}[1]{\macro{\weightParamLetter_{\mathrm{#1}}}}
\newcommand{\weightParamL}[2]{\macro{\weightParamLetter^{(#2)}_{\mathrm{#1}}}}
\newcommand{\biasParamLetter}{\macro{\vecVar{b}}}
\newcommand{\biasParam}[1]{\macro{\biasParamLetter_{\mathrm{#1}}}}
\newcommand{\biasParamL}[2]{\macro{\biasParamLetter^{(#2)}_{\mathrm{#1}}}}
\newcommand{\dropout}[1]{\textsc{Dropout}(#1)}
\newcommand{\layerNorm}[1]{\textsc{LayerNorm}(#1)}
\newcommand{\softmax}{\mathrm{softmax}}

\newcommand{\theStackVectorSize}{\macro{m}}
\newcommand{\theStackObjectT}[1]{\macro{\mathcal{S}}_{#1}}
\newcommand{\theStackActionVectorT}[1]{\macro{\vecVar{a}}_{#1}}
\newcommand{\theStackActionLogitsT}[1]{\vecVar{a}'_{#1}}
\newcommand{\theNumActions}{\macro{a}}
\newcommand{\theStackReadingSize}{\macro{r}}
\newcommand{\thePushedVectorT}[1]{\macro{\vecVar{v}}_{#1}}
\newcommand{\theStackReadingT}[1]{\macro{\vecVar{r}}_{#1}}
\newcommand{\stackFunctionName}{\macro{\textsc{Stack}}}
\newcommand{\readingFunctionName}{\macro{\textsc{Reading}}}
\newcommand{\actionFunctionName}{\macro{\textsc{Actions}}}
\newcommand{\stackFunction}[3]{\stackFunctionName(#1, #2, #3)}
\newcommand{\readingFunction}[1]{\readingFunctionName(#1)}
\newcommand{\actionFunction}[1]{\actionFunctionName(#1)}
\newcommand{\theRun}{\macro{\bm{\pi}}}
\newcommand{\topVectorOfRun}[1]{\vecVar{v}(#1)}
\newcommand{\weightOfRun}[1]{\psi(#1)}

\newcommand{\theSupStackTensorT}[1]{\matVar{V}_{#1}}
\newcommand{\theSupStackTensorTI}[2]{(\theSupStackTensorT{#1})_{#2}}
\newcommand{\theStackPos}{\macro{i}}
\newcommand{\pushLabel}{\macro{\textsc{push}}}
\newcommand{\noopLabel}{\macro{\textsc{noop}}}
\newcommand{\popLabel}{\macro{\textsc{pop}}}
\newcommand{\theSupActionProbT}[2]{a_{#1}^{#2}}
\newcommand{\theSupPushProbT}[1]{\theSupActionProbT{#1}{\pushLabel}}
\newcommand{\theSupNoopProbT}[1]{\theSupActionProbT{#1}{\noopLabel}}
\newcommand{\theSupPopProbT}[1]{\theSupActionProbT{#1}{\popLabel}}
\newcommand{\theStackActionT}[1]{\tau_{#1}}
\newcommand{\supRunSet}[1]{\Pi[\pushLabel \leadsto #1]}

\newcommand{\theStateSet}{\macro{Q}}
\newcommand{\theStackAlphabet}{\macro{\Gamma}}
\newcommand{\pdaTransition}[5]{#1, #3 \xrightarrow{#2} #4, #5}
\newcommand{\theStateFrom}{\macro{q}}
\newcommand{\theStateTo}{\macro{r}}
\newcommand{\stackBottomSymbol}{\macro{\bot}}
\newcommand{\theInitialState}{\macro{q_0}}
\newcommand{\thePoppedSymbol}{\macro{x}}
\newcommand{\thePushedSymbol}{\macro{y}}
\newcommand{\thePoppedVector}{\macro{\vecVar{u}}}

\newcommand{\theAlphabet}{\macro{\Sigma}}
\newcommand{\theSymbol}{\macro{b}}
\newcommand{\theTimestep}{\macro{t}}
\newcommand{\theOtherTimestep}{\macro{i}}
\newcommand{\theOtherOtherTimestep}{\macro{j}}
\newcommand{\theString}{\macro{\stringVar{w}}}
\newcommand{\theSymbolT}[1]{w_{#1}}
\newcommand{\theLength}{\macro{n}}
\newcommand{\theTransition}{\macro{\tau}}
\newcommand{\theTransitionT}[1]{\tau_{#1}}
\newcommand{\weightOfTransition}[1]{\psi(#1)}
\newcommand{\theFinalState}{\macro{r}}
\newcommand{\theTopStackSymbol}{\macro{y}}
\newcommand{\ndRunSet}[3]{\Pi[\theInitialState, \stackBottomSymbol, 0 \leadsto #2, #3, #1]}
\newcommand{\theNDStackReadingSlice}[3]{\vecVar{r}_{#1}[#2, #3]}

\newcommand{\theModel}{\macro{M}}
\newcommand{\theModelProbDist}{\macro{p_{\theModel}}}
\newcommand{\theLayerNo}{\macro{\ell}}
\newcommand{\theNumLayers}{\macro{L}}
\newcommand{\theModelSize}{\macro{d}}
\newcommand{\dmodel}{\macro{d_\mathrm{model}}}
\newcommand{\theInputEmbeddingMatrix}{\macro{\matVar{E}}}
\newcommand{\theInputEmbeddingT}[1]{\macro{\vecVar{x}_{#1}}}

\newcommand{\theLogitsT}[1]{\macro{\vecVar{y}}_{#1}}

\newcommand{\theDataset}{\macro{D}}
\newcommand{\theInputString}{\macro{\stringVar{x}}}
\newcommand{\theOutputString}{\macro{\stringVar{y}}}
\newcommand{\theBatchSize}{\macro{B}}
\newcommand{\fullAccuracy}[2]{\macro{\mathrm{FA}}(#1, #2)}
\newcommand{\theConditionalProbability}{\macro{\theModelProbDist(\theOutputString \mid \theInputString)}}

\newcommand{\theRNNInputT}[1]{\vecVar{x}'_{#1}}
\newcommand{\theRNNStateT}[1]{\mathcal{H}_{#1}}
\newcommand{\theHiddenStateLT}[2]{\macro{\vecVar{h}}^{(#1)}_{#2}}
\newcommand{\theRNNOutputT}[1]{\vecVar{y}'_{#1}}
\newcommand{\theHiddenStateDropoutLT}[2]{\macro{\cancel{\vecVar{h}}^{(#1)}_{#2}}}
\newcommand{\theInitialHiddenStateParamL}[1]{\macro{\vectorParamL{w}{0}{#1}}}
\newcommand{\theInputGateLT}[2]{\macro{\vecVar{i}^{(#1)}_{#2}}}
\newcommand{\theForgetGateLT}[2]{\macro{\vecVar{f}^{(#1)}_{#2}}}
\newcommand{\theCandidateLT}[2]{\macro{\vecVar{g}^{(#1)}_{#2}}}
\newcommand{\theOutputGateLT}[2]{\macro{\vecVar{o}^{(#1)}_{#2}}}
\newcommand{\theMemoryCellLT}[2]{\macro{\vecVar{c}^{(#1)}_{#2}}}
\newcommand{\recurrenceFunctionName}{\macro{\textsc{Recurrence}}}
\newcommand{\recurrenceFunction}[2]{\recurrenceFunctionName(#1, #2)}

\newcommand{\theTfSublayerInputT}[1]{\bar{\vecVar{x}}_{#1}}
\newcommand{\theTfSublayerLNOutputT}[1]{\bar{\vecVar{x}}^\mathrm{LN}_{#1}}
\newcommand{\theTfSublayerFuncOutputT}[1]{\bar{\vecVar{y}}^\mathrm{SL}_{#1}}
\newcommand{\theTfSublayerOutputT}[1]{\bar{\vecVar{y}}_{#1}}
\newcommand{\sublayerFunctionName}{\macro{\textsc{Sublayer}}}

\newcommand{\theGeneralizationSet}{\macro{\theDataset_\mathrm{hier}}}
\newcommand{\theLinearGeneralizationSet}{\macro{\theDataset_\mathrm{lin}}}

\title{Bearing Syntactic Fruit with Stack-Augmented Neural Networks}

\author{Brian DuSell \\
  ETH Z{\"u}rich \\
  \href{mailto:brian.dusell@inf.ethz.ch}{\texttt{brian.dusell@inf.ethz.ch}} \\\And
  Ryan Cotterell \\
  ETH Z{\"u}rich \\
  \href{mailto:ryan.cotterell@inf.ethz.ch}{\texttt{ryan.cotterell@inf.ethz.ch}} \\}

\begin{document}
\maketitle
\begin{abstract}
When children learn language, they make syntactic generalizations based on hierarchical rules.
A recent line of work has inquired as to whether common neural network architectures share this inductive bias for hierarchical syntax, finding that they do so only under special conditions: when augmented with ground-truth parse tree structures, when pre-trained on massive corpora, or when trained long past convergence. In this paper, we demonstrate, for the first time, neural network architectures that generalize in human-like fashion when trained only on surface forms: stack-augmented neural networks. We test three base architectures (transformer, simple RNN, LSTM) augmented with two styles of stack, one of which leverages nondeterminism. We find that transformers with nondeterministic stacks generalize best on multiple tasks designed to measure hierarchical inductive bias.
This suggests that stack-augmented neural networks may be more accurate models of human syntax acquisition than standard architectures, serving as useful objects of psycholinguistic study. Our code \ifAnonymousSubmissionMode{}will be\else{}is\fi{} publicly available.\footnote{\ifAnonymousSubmissionMode{}See the anonymous supplementary material.\else{}\url{https://github.com/bdusell/bearing-syntactic-fruit}\fi}
\end{abstract}

\section{Introduction}

\def\customIndent{2em}

When observing a finite set of input-output examples for a task, a learner must anticipate unseen inputs by inferring a plausible algorithm that could have generated the outputs from the inputs. Multiple algorithms are always possible, and the learner must choose one of them. Consider, for instance, the task of \newterm{question formation}: converting a declarative sentence to its equivalent question form.
\begin{itemize}[label={},leftmargin=\customIndent,topsep=0.1em,itemsep=0.05em]
    \item The salamanders don't amuse my newt. \\
        $\mapsto$ Don't the salamanders amuse my newt?
    \item My walrus does move. \\
        $\mapsto$ Does my walrus move?
\end{itemize}
The examples above are compatible with the following two algorithms (at least).
\newenvironment{inlinefigure}
    {\begin{adjustwidth}{\customIndent}{\customIndent}}
    {\end{adjustwidth}}
\vspace{0.5em}
\begin{inlinefigure}
    \moveMain{}: Parse the sentence into a noun phrase (NP) and verb phrase (VP); front the auxiliary verb at the beginning of the VP.
    \begin{center}
        \vspace{-0.5em}
        \begin{forest}
            for tree={
                align=center,
                baseline,
                edge=-,
                inner sep=0pt,
                minimum height=0pt,
                s sep=0.25cm,
                l=0.1cm,
                font=\footnotesize
            }
            [ [ NP [the,name=firstword] [salamanders] ] [ VP [\textDelete{don't},name=deletedword] [ [amuse] [ [my] [newt\textDelete{.}\textInsert{?}] ] ] ] ]
            \node(insertedword)[left=0.2cm of firstword.base west, anchor=base east] {\footnotesize \textInsert{Don't}};
            \draw[->,>=Stealth,bend left=45,shorten >=5pt,shorten <=5pt] (deletedword.base) to (insertedword.base);
        \end{forest}
    \end{center}
\end{inlinefigure}
\begin{inlinefigure}
    \moveFirst{}: Front the first auxiliary verb.
    \begin{center}
        \vspace{-2.8em}
        \begin{tikzpicture}[
            baseline=(current bounding box.north),
            every node/.style={font=\footnotesize}
        ]
            \def\s{-4pt}
            \node (D1) {\textInsert{Don't}};
            \node (the) [right=\s of D1.base east, anchor=base west] {the};
            \node (salamanders) [right=\s of the.base east, anchor=base west] {salamanders};
            \node (D2) [right=\s of salamanders.base east, anchor=base west] {\textDelete{don't}};
            \node (amuse) [right=\s of D2.base east, anchor=base west] {amuse};
            \node (my) [right=\s of amuse.base east, anchor=base west] {my};
            \node (newt) [right=\s of my.base east, anchor=base west] {newt\textDelete{.}\textInsert{?}};
            \draw[->,>=Stealth,bend left=35,shorten >=5pt,shorten <=5pt] (D2.base) to (D1.base);
        \end{tikzpicture}
    \end{center}
\end{inlinefigure}
The \moveMain{} algorithm, which is the linguistically correct choice, syntactically parses the sentence to identify the main auxiliary verb, whereas \moveFirst{} incorrectly relies on linear position alone. On the examples above, both algorithms produce the same result; they would only differ on an input like the following.

\begin{inlinefigure}
    The salamanders who do sleep don't amuse my newt.
    \begin{itemize}[label={},leftmargin=0em,topsep=0.05em,itemsep=0.05em]
        \item $\mapsto$ Don't the salamanders who do sleep amuse my newt? (\moveMain{})
        \item $\mapsto$ Do the salamanders who sleep don't amuse my newt? (\moveFirst{})
    \end{itemize}
\end{inlinefigure}
Here, \moveMain{} correctly groups ``The salamanders who do sleep'' into a single NP and fronts \textit{don't}, whereas \moveFirst{} incorrectly fronts the \textit{do} from inside the relative clause. Evaluating a learner on such an input would reveal its preference for \moveMain{}, \moveFirst{}, or some other algorithm; we call this preference the learner's \newterm{inductive bias}.

The \moveMain{} algorithm requires an assumption that language conforms to hierarchical syntax, whereas \moveFirst{} simply uses linear position.
Linguistic studies have shown that human children consistently form questions according to \moveMain{} \citep{crain-nakayama-1987-structure}, even though they have probably encountered very few disambiguating examples like the above (though some, like \citet{pullum-scholz-2002-empirical}, have disputed this latter claim). Chomsky argued that this phenomenon reveals an inherent bias of human learners for hierarchical syntax---the so-called ``argument from the poverty of the stimulus'' \citeyearpar{chomsky-1965-aspects,chomsky-1980-rules}.

\Citet{mccoy-etal-2020-syntax} investigated whether commonly used neural network architectures exhibit a similar bias. Their work and several follow-ups \citep{mulligan-etal-2021-structure,petty-frank-2021-transformers,mueller-etal-2022-coloring,mueller-linzen-2023-plant,yedetore-etal-2023-poor,murty-etal-2023-grokking,mueller-etal-2024-context,yedetore-kim-2024-semantic,ginn-2024-tree,ahuja-etal-2025-learning,qin-etal-2025-data} have answered this largely in the negative, except under special circumstances: namely, when a model is augmented with ground-truth parse trees \citep{mccoy-etal-2020-syntax}, pre-trained on
large corpora \citep{mueller-etal-2022-coloring,yedetore-etal-2023-poor}, trained on semantic representations \citep{yedetore-kim-2024-semantic}, or trained long after validation performance converges \citep{murty-etal-2023-grokking,ahuja-etal-2025-learning,qin-etal-2025-data}.
It has remained an open question whether there is a neural network architecture capable of generalizing in human-like fashion without requiring such strong syntactic supervision; as \citet{mccoy-etal-2020-syntax} put it, ``Does syntax need to grow on trees?''

In this paper, we explore this question on an architecture previously unstudied in this setting: stack-augmented neural networks. Stack-augmented neural networks are sequential models that consist of a standard base architecture (e.g., an RNN or transformer) connected to a differentiable stack data structure \citep{joulin-mikolov-2015-inferring,dusell-chiang-2022-learning,dusell-chiang-2023-surprising,dusell-chiang-2024-stack}. Stacks are a cornerstone of context-free parsing algorithms and so are a natural choice for imbuing architectures with a bias for hierarchical syntax. We show that some of these stack-augmented architectures do, in fact, show a clear preference for hierarchical generalization on the question formation task when trained simply on surface forms, often fronting not only the correct verb, but generating the entire output in accordance with \moveMain{} (in one case the model attains an average accuracy of 32\%, in contrast with a typical accuracy of 0\% in prior work).
We show improved hierarchical generalization for other tasks in English and German as well. In this way, we show that one can ``bear syntactic fruit'' without explicit parse trees.

Our findings contribute to a broader debate about whether the acquisition of hierarchical generalization emanates from the learner (i.e., physiological biases in the human brain, as \citet{elman-etal-1996-rethinking} suggested) or from the stimulus. In a sense, a learning algorithm's bias always indicates a property of the \emph{learner}; for any training set, one can adversarially construct a learning algorithm that ignores any hints one way or the other. Moreover, the notion of ``simplicity'' is always relative to the learner's parameterization. The question, then, is really whether a reasonably simple learning algorithm---not the kind of contrived example just mentioned, but perhaps a neural network architecture with minimal assumptions about the specific task---can learn a rule like \moveMain{} from ambiguous data while still attaining competitive performance on natural language benchmarks. We offer stack-augmented neural networks as a positive example that hierarchical generalization can emerge from surface forms alone given the right neural network architecture.

\section{Testing the Poverty of the Stimulus}
\label{sec:tasks}

The purpose of our experiments is to test whether a given neural network architecture, when trained on data that ambiguously conforms to rules based on hierarchical syntax and linear position, generalizes in accordance with the hierarchical rule. We use an established experimental framework to do so \citep{wilson-2006-learning,frank-mathis-2007-transformational,mccoy-etal-2018-revisiting}, specifically using the \newterm{question formation} task of \citet{mccoy-etal-2020-syntax}, the \newterm{German question formation} task of \citet{mueller-etal-2022-coloring}, the \newterm{tense reinflection with \textit{do}} task provided alongside the code of \citet{mccoy-etal-2020-syntax}, and the \newterm{passivization} task of \citet{mueller-etal-2022-coloring}. All tasks involve transforming an input sentence to an output sentence. All input sentences were sampled from a simple probabilistic context-free grammar with a small vocabulary. Each task has a \newterm{training set} of 100k examples, a \newterm{validation set} of 1k examples, and an in-distribution \newterm{test set} of 10k examples where all examples are consistent with both the hierarchical and linear rule. Each task also has an out-of-distribution \newterm{generalization set} of 10k examples that is consistent only with the hierarchical rule; evaluating on this dataset reveals the model's inductive bias.

In the question formation task (\cref{fig:question-formation-examples}), the input sentence always ends in a special token that indicates one of two types of transformation to apply: \qfDecl{}, which means to copy the declarative sentence to the output unchanged; and \qfQuest{}, which means to convert it to a question. In examples with \qfDecl{}, the subject noun phrase may or may not be followed by a relative clause containing a verb, so the main verb is not necessarily the first verb in the sentence.
On the other hand, in examples with \qfQuest{}, the subject noun phrase is never followed by a relative clause containing a verb, so the main verb always happens to be the first verb in the sentence. Therefore, the training data is consistent with both \moveMain{} and \moveFirst{}. The generalization set consists entirely of examples that contain \qfQuest{} where the subject noun phrase is followed by a relative clause with a verb, in which case only \moveMain{} correctly predicts the output.

\begin{figure*}[t]
    \centering
    \begin{subfigure}{\linewidth}
        \centering
        \small
        \begin{tabular}{lcl}
            my raven does change . \qfDecl{} & $\mapsto$ & my raven does change . \\
            my raven does change . \qfQuest{} & $\mapsto$ & does my raven change ? \\
            \rowcolor{lightgray} my raven that doesn't sleep does change . \qfQuest{} & $\mapsto$ & does my raven that doesn't sleep change ? (\moveMain{}) \\
            \rowcolor{lightgray} & $\mapsto$ & doesn't my raven that sleep does change ? (\moveFirst{})
        \end{tabular}
        \caption{Question Formation}
        \label{fig:question-formation-examples}
    \end{subfigure}
    \begin{subfigure}{\linewidth}
        \centering
        \small
        \begin{tabular}{lcl}
            dein Rabe kann feiern . \qfDecl{} & $\mapsto$ & dein Rabe kann feiern . \\
            dein Rabe kann feiern . \qfQuest{} & $\mapsto$ & kann dein Rabe feiern ? \\
            \rowcolor{lightgray} dein Rabe , der gegessen hat , kann feiern . \qfQuest{} & $\mapsto$ & kann dein Rabe , der gegessen hat , feiern ? (\moveMain{}) \\
            \rowcolor{lightgray} & $\mapsto$ & hat dein Rabe , der gegessen , kann feiern ? (\moveFirst{})
        \end{tabular}
        \caption{German Question Formation}
        \label{fig:question-formation-in-german-examples}
    \end{subfigure}
    \begin{subfigure}{\linewidth}
        \centering
        \small
        \begin{tabular}{lcl}
            some newts did smile . \trPast{} & $\mapsto$ & some newts did smile . \\
            some newts did smile . \trPresent{} & $\mapsto$ & some newts do smile . \\
            \rowcolor{lightgray} some newts by our yak did smile . \trPresent{} & $\mapsto$ & some newts by our yak do smile . (\agreeMain{}) \\
            \rowcolor{lightgray} & $\mapsto$ & some newts by our yak does smile . (\agreeRecent{})
        \end{tabular}
        \caption{Tense Reinflection with \textit{Do}}
        \label{fig:tense-reinflection-with-do-examples}
    \end{subfigure}
    \begin{subfigure}{\linewidth}
        \centering
        \small
        \begin{tabular}{lcl}
            her yak amused her newt . \passDecl{} & $\mapsto$ & her yak amused her newt . \\
            her yak amused her newt . \passPassiv{} & $\mapsto$ & her newt was amused by her yak . \\
            \rowcolor{lightgray} her yak by her zebras amused her newt . \passPassiv{} & $\mapsto$ & her newt was amused by her yak by her zebras . (\moveObject{}) \\
            \rowcolor{lightgray} & $\mapsto$ & her zebras were amused by her yak . (\moveSecond{})
        \end{tabular}
        \caption{Passivization}
        \label{fig:passivization-examples}
    \end{subfigure}
    \caption{Examples that illustrate the difference between the training (shaded white) and generalization (shaded gray) sets of each task.}
    \label{fig:task-examples}
\end{figure*}

We include experiments on three other similarly designed tasks.
The German question formation task \citep{mueller-etal-2022-coloring} entails transforming a declarative sentence in German to its equivalent question form (\cref{fig:question-formation-in-german-examples}). It is structurally very similar to English question formation, and the competing rules are again \moveMain{} and \moveFirst{}. The tense reinflection with \textit{do} task entails transforming all past-tense verbs in a sentence to present tense, inflecting them to agree in number with their subjects (\cref{fig:tense-reinflection-with-do-examples}). The hierarchical rule, \agreeMain{}, inflects verbs to agree with their grammatical subjects, whereas the linear rule, \agreeRecent{}, inflects each verb to agree with its most recent noun. This task is based on the tense reinflection task of \citet{mccoy-etal-2020-syntax}, but instead of using inflected forms unique to each verb type, all verbs use the auxiliary \textit{do}. We chose this variant of the task because it is simpler to learn only the inflected forms for \textit{do}. Finally, passivization \citep{mueller-etal-2022-coloring} entails converting a sentence in active voice to the equivalent sentence in passive voice, swapping the subject and object (\cref{fig:passivization-examples}). The hierarchical rule, \moveObject{}, moves the object noun phrase to subject position, whereas the linear rule, \moveSecond{}, moves the second noun phrase.\footnote{We also include results on the modified question formation task of \citet{mueller-etal-2022-coloring}, the original tense reinflection task of \citet{mccoy-etal-2020-syntax}, and the German passivization task of \citet{mueller-etal-2022-coloring} in \cref{sec:full-results}. We omit German passivization from our main results due to issues in the task design; see \cref{sec:passivization-in-german-issues}.}

\section{Differentiable Stacks}

We now review stack-augmented neural network architectures. Stacks are a central component of many context-free grammar parsing algorithms, as their last-in-first-out protocol makes them suited for tracking unclosed constituents (in top-down parsing) or completed constituents (in bottom-up parsing) in the proper order. Stack-augmented neural network architectures consist of a standard base architecture (i.e., a simple RNN, LSTM, or transformer) connected to a \newterm{differentiable stack}. We first explain differentiable stacks and later explain how they interface with the base architecture.

A differentiable stack is a continuous function that simulates the behavior of a discrete stack; there are multiple kinds, which we will discuss below. Conceptually, a differentiable stack converts the familiar operations of pushing, popping, and reading the top element to continuous relaxations. Instead of receiving a single push or pop command at a time, it receives a set of weighted \newterm{stack actions}, which it simulates in proportion to their weights. The differentiable stack then produces a \newterm{stack reading} that represents an interpolation of the topmost stack element after applying those action weights. A neural network can compute the stack actions dynamically in one part and use the stack reading as input in another part. Because the stack reading is differentiable with respect to the stack actions, the whole network can still be trained end-to-end with ordinary backpropagation and gradient descent---and without needing supervision over the stack actions. In this way, the differentiable stack provides the model with a latent representation of syntactic structure.

For both types of differentiable stack that we use in this paper, the differentiable stack simulates a discrete stack whose elements are vectors in $(0, 1)^{\theStackVectorSize}$, where $\theStackVectorSize$ is a hyperparameter. The differentiable stack starts in an initial, empty state $\theStackObjectT{0}$. We can iteratively apply actions to the differentiable stack and extract a stack reading at each step. At iteration $\theTimestep > 0$, let $\theStackActionVectorT{\theTimestep} \in \positiveRealSet^{\theNumActions}$ be a vector of stack action weights, let $\theStackActionLogitsT{\theTimestep} \in \realSet^{\theNumActions}$ be a vector of unnormalized logits used to compute $\theStackActionVectorT{\theTimestep}$, let $\thePushedVectorT{\theTimestep} \in (0, 1)^{\theStackVectorSize}$ be the vector pushed to the stack by the push action, and let $\theStackReadingT{\theTimestep} \in \positiveRealSet^{\theStackReadingSize}$ be the resulting stack reading. We can abstract the differentiable stack into three functions: $\actionFunctionName$, which converts $\theStackActionLogitsT{\theTimestep}$ to $\theStackActionVectorT{\theTimestep}$; $\stackFunctionName$, which incrementally updates the stack; and $\readingFunctionName$, which produces the stack reading.
\begin{align}
    \theStackActionVectorT{\theTimestep} &\defeq \actionFunction{\theStackActionLogitsT{\theTimestep}} \\
    \theStackObjectT{\theTimestep} &\defeq \stackFunction{\theStackObjectT{\theTimestep-1}}{\theStackActionVectorT{\theTimestep}}{\thePushedVectorT{\theTimestep}} \\
    \theStackReadingT{\theTimestep} &\defeq \readingFunction{\theStackObjectT{\theTimestep}}
\end{align}

\subsection{Superposition Stack}
\label{sec:superposition-stack}

The \newterm{superposition stack} of \citet{joulin-mikolov-2015-inferring} assumes that the action vector is a probability distribution over three actions: push, no-op, and pop. We have $\theStackActionVectorT{\theTimestep} \defeq (\theSupPushProbT{\theTimestep}, \theSupNoopProbT{\theTimestep}, \theSupPopProbT{\theTimestep}) \defeq \softmax(\theStackActionLogitsT{\theTimestep})$. The stack $\theStackObjectT{\theTimestep}$ is a matrix $\theSupStackTensorT{\theTimestep} \in \realSet^{(\theTimestep+1) \times \theStackVectorSize}$, where the vector $\theSupStackTensorTI{\theTimestep}{\theStackPos}$ represents the $\theStackPos$\textsuperscript{th} element from the top of the stack. The initial stack, $\theSupStackTensorT{0}$, contains a single $\zeroVector$ vector. The function $\stackFunctionName$ works by computing three new, separate stacks: one with all elements shifted down and $\thePushedVectorT{\theTimestep}$ inserted at the top (push); one kept the same (no-op); and one with all elements shifted up and the topmost deleted (pop). The new stack is formed by interpolating these stacks elementwise according to the push, no-op, and pop probabilities, making a superposition of the three. The function $\readingFunctionName$ simply returns the top vector in $\theSupStackTensorT{\theTimestep}$. See \cref{sec:superposition-stack-details} and \citet{dusell-chiang-2024-stack} for details.

\subsection{Nondeterministic Stack}

The \newterm{nondeterministic stack} of \citet{dusell-chiang-2020-learning,dusell-chiang-2022-learning,dusell-chiang-2023-surprising} simulates a nondeterministic pushdown automaton (PDA) and is a generalization of the superposition stack.
More precisely, we use what \citet{dusell-chiang-2024-stack} call the \newterm{differentiable vector PDA (dVPDA)}, which is a continuous function that simulates a slightly modified PDA called a \newterm{vector PDA (VPDA)}.
A standard PDA has a finite set of states $\theStateSet$, a stack data structure whose elements come from a finite alphabet $\theStackAlphabet$, and a finite set of transitions, which dictate how the state machine and stack interact.
In a VPDA, elements of the stack are members of $\theStackAlphabet \times (0, 1)^{\theStackVectorSize}$. The two parts of these stack elements serve different roles: the discrete symbol from $\theStackAlphabet$ interacts with the PDA's state machine, whereas the vector from $(0, 1)^{\theStackVectorSize}$ serves as an efficient way to encode information tacked onto each stack element.

Let $\theString = \theSymbolT{1} \cdots \theSymbolT{\theLength}$ be a string of input symbols. The VPDA starts in an initial state $\theInitialState \in \theStateSet$ with a stack that contains only $(\stackBottomSymbol, \thePushedVectorT{0})$, where $\stackBottomSymbol \in \theStackAlphabet$ is a designated bottom symbol, and $\thePushedVectorT{0}$ is a learned initial bottom vector. The VPDA has three types of transition, where $\theStateFrom, \theStateTo \in \theStateSet$ and $\thePoppedSymbol, \thePushedSymbol \in \theStackAlphabet$, with the following semantics.
\begin{itemize}[itemsep=0.05em]
    \item $\pdaTransition{\theStateFrom}{\theSymbolT{\theTimestep}}{\thePoppedSymbol}{\theStateTo}{\thePoppedSymbol\thePushedSymbol}$: If $(\thePoppedSymbol, \thePoppedVector)$ is on top, \newterm{push} $(\thePushedSymbol, \thePushedVectorT{\theTimestep})$.
    \item $\pdaTransition{\theStateFrom}{\theSymbolT{\theTimestep}}{\thePoppedSymbol}{\theStateTo}{\thePushedSymbol}$: If $(\thePoppedSymbol, \thePoppedVector)$ is on top, \newterm{replace} it with $(\thePushedSymbol, \thePoppedVector)$.
    \item $\pdaTransition{\theStateFrom}{\theSymbolT{\theTimestep}}{\thePoppedSymbol}{\theStateTo}{\emptyString}$: If $(\thePoppedSymbol, \thePoppedVector)$ is on top, \newterm{pop} it.
\end{itemize}

A \emph{deterministic} VPDA would only admit at most one outgoing transition for each state $\theStateFrom$ and top stack symbol $\thePoppedSymbol$. In contrast, the dVPDA simulates a \emph{nondeterministic} VPDA that allows multiple, in which case it simulates all possible sequences of transitions. In other words, the dVPDA does not have to commit to a single parse of its input while incrementally processing it. Rather, it can encode a \emph{distribution} over all possible syntactic structures. See \cref{sec:nondeterministic-stack-details} and \citet{dusell-chiang-2023-surprising} for details.

\section{Stack-Augmented Neural Networks}
\label{sec:stack-nn-interface}

Here, we describe how to combine differentiable stacks with three base architectures: the simple RNN \citep{elman-1990-finding}, LSTM \citep{hochreiter-schmidhuber-1997-long}, and transformer \citep{vaswani-etal-2017-attention}. For simplicity, we will use one free hyperparameter $\theModelSize$ to control the size of the model and express other hyperparameters in terms of $\theModelSize$. For the RNN and LSTM, $\theModelSize$ is the number of hidden units; for the transformer, $\theModelSize$ is the model width $\dmodel$. Let $\theAlphabet$ be a finite vocabulary of tokens, and let $\theString = \theSymbolT{1} \cdots \theSymbolT{\theLength} \in \theAlphabet^{\theLength}$ be the input string. Assume the model is a language model that computes logit vectors $\theLogitsT{0}, \ldots, \theLogitsT{\theLength} \in \realSet^{|\theAlphabet|+1}$, where for each $0 \leq \theTimestep \leq \theLength$, $\theSymbol \in \theAlphabet \cup \{\eos\}$, the next-token probability is $\theModelProbDist(\theSymbol \mid \theSymbolT{1} \cdots \theSymbolT{\theTimestep}) \defeq \softmax(\theLogitsT{\theTimestep})_{\theSymbol}$. In the following, $\logistic{\cdot}$ is the logistic function, and $\dropout{\cdot}$ indicates the application of dropout \citep{srivastava-etal-2014-dropout}.

\subsection{Stack RNN and LSTM}

Expressivity and regularization play important roles in hierarchical generalization; for this reason, we describe not only how to connect RNNs and LSTMs to differentiable stacks, but also three new modifications of ours meant to promote generalization: we use multiple layers in the base RNN or LSTM, we add dropout, and we change how the stack reading is used. First, we describe the base architectures. Let $\theNumLayers$ be the number of layers. At each step $\theTimestep$, the model computes a series of hidden states $\theHiddenStateLT{1}{\theTimestep}, \ldots, \theHiddenStateLT{\theNumLayers}{\theTimestep}$. (In the case of the LSTM, it also computes a series of memory cell values.) Let us encapsulate the entire state of the model at step $\theTimestep$ in a single object $\theRNNStateT{\theTimestep}$, and let $\recurrenceFunctionName$ be the function that updates the state. Each architecture has an initial state $\theRNNStateT{0}$. Let $\theInputEmbeddingMatrix \in \realSet^{(|\theAlphabet|+1) \times \theModelSize}$ be a learned embedding matrix, and let $\theInputEmbeddingT{\theTimestep} \defeq \theInputEmbeddingMatrix_{\theSymbolT{\theTimestep}}$ be the input embedding of $\theSymbolT{\theTimestep}$. We apply dropout between layers like \citet{zaremba-etal-2015-recurrent}, and we tie the input and output embeddings \citep{press-wolf-2017-using}. For $0 < \theTimestep \leq \theLength$, $\theRNNInputT{\theTimestep} \defeq \dropout{\theInputEmbeddingT{\theTimestep}}$, $\theRNNStateT{\theTimestep} \defeq \recurrenceFunction{\theRNNStateT{\theTimestep-1}}{\theRNNInputT{\theTimestep}}$, $\theRNNOutputT{\theTimestep} \defeq \dropout{\theHiddenStateLT{\theNumLayers}{\theTimestep}}$, and $\theLogitsT{\theTimestep} \defeq \theInputEmbeddingMatrix \theRNNOutputT{\theTimestep}$. See \cref{sec:rnn-details,sec:lstm-details} for details.

To connect a base RNN or LSTM, called the \newterm{controller}, to a differentiable stack, we use the last layer's hidden state to compute the action logits as $\theStackActionLogitsT{\theTimestep} \defeq \affine{a}{\theHiddenStateLT{\theNumLayers}{\theTimestep}}$ and the pushed vector as $\thePushedVectorT{\theTimestep} \defeq \logistic{\affine{v}{\theHiddenStateLT{\theNumLayers}{\theTimestep}}}$, where $\weightParam{a} \in \realSet^{\theNumActions \times \theModelSize}$, $\biasParam{a} \in \realSet^{\theNumActions}$, $\weightParam{v} \in \realSet^{\theStackVectorSize \times \theModelSize}$, and $\biasParam{v} \in \realSet^{\theStackVectorSize}$ are learned parameters. We then use the stack reading as an extra input to the next update, giving $\theRNNStateT{\theTimestep} \defeq \recurrenceFunction{\theRNNStateT{\theTimestep-1}}{\begin{bmatrix}
    \theRNNInputT{\theTimestep} \\
    \theStackReadingT{\theTimestep}
\end{bmatrix}}$. See \cref{fig:stack-rnn-diagram} for an illustration.

\begin{figure}[t]
    \centering
    \scalebox{0.8}{\begin{tikzpicture}[x=1cm,y=1cm,>=Stealth]
    \def\ysep{2}
    \def\stacky{-1.1}
    \def\inputy{-2.4}
    \def\xstep{2}
    \def\leftdotsx{-1.7*\xstep}
    \def\stackleftdotsx{-1.5*\xstep}
    \def\labelx{-1.8*\xstep}
    \newcommand{\dropoutLabel}{\scalebox{0.7}{$\funcName{Dropout}$}}
    \newcommand{\stepNodes}[3]{%
        \node (x #2) at (#1, \inputy) {$\theInputEmbeddingT{#3}$};
        \node (dr0 #2) at (#1, \inputy+0.75) {\dropoutLabel};
        \node (h1 #2) at (#1, 0) {$\theHiddenStateLT{1}{#3}$};
        \node (dr1 #2) at (#1, 0.5*\ysep) {\dropoutLabel};
        \node (h2 #2) at (#1, \ysep) {$\theHiddenStateLT{2}{#3}$};
        \node (dr2 #2) at (#1, 1.5*\ysep) {\dropoutLabel};
        \node (y #2) at (#1, 2*\ysep) {$\theLogitsT{#3}$};
    }
    \newcommand{\stepEdges}[4]{%
        \draw[->] (x #1) edge (dr0 #1);
        \draw[->] (dr0 #1) edge (h1 #1);
        \draw[->] (h1 #1) edge (dr1 #1);
        \draw[->] (dr1 #1) edge (h2 #1);
        \draw[->] (h2 #1) edge (dr2 #1);
        \draw[->] (dr2 #1) edge (y #1);
        \draw[->] (s #2) edge node[sloped,above] {$\theStackReadingT{#4}$} (h1 #1);
        \draw[->] (h2 #1) edge node[sloped,above] {$\theStackActionVectorT{#3}, \thePushedVectorT{#3}$} (s #1);
        \draw[->] (s #2) edge (s #1);
        \draw[->, dashed, bend right] (s #1) edge (dr2 #1);
    }
    \newcommand{\betweenNodes}[3]{%
        \node (s #2) at (#1, \stacky) {$\theStackObjectT{#3}$};
    }
    \newcommand{\betweenEdges}[2]{%
        \draw[->] (h1 #1) edge (h1 #2);
        \draw[->] (h2 #1) edge (h2 #2);
    }
    \newcommand{\stepDotsNodes}[2]{%
        \node (h1 #2) at (#1, 0) {$\cdots$};
        \node (h2 #2) at (#1, \ysep) {$\cdots$};
    }

    \stepDotsNodes{\leftdotsx}{t-2}
    \node (s t-2) at (\stackleftdotsx, \stacky) {$\cdots$};
    \stepNodes{-1*\xstep}{t-1}{\theTimestep-1}
    \betweenNodes{-0.5*\xstep}{t-1}{\theTimestep-1}
    \stepNodes{0*\xstep}{t}{\theTimestep}
    \betweenNodes{0.5*\xstep}{t}{\theTimestep}
    \stepNodes{1*\xstep}{t+1}{\theTimestep+1}
    \node (s t+1) at (1.5*\xstep, \stacky) {$\cdots$};
    \stepDotsNodes{2*\xstep}{t+2}

    \node[anchor=east] at (\labelx,2*\ysep) {output \big\{};
    \node[anchor=east] at (\labelx,0.5*\ysep) {$\big\{$};
    \node[anchor=south, rotate=90, yshift=10pt] at (\labelx,0.5*\ysep) {controller};
    \node[anchor=east] at (\labelx,\stacky) {stack \big\{};
    \node[anchor=east] at (\labelx,\inputy) {input \big\{};

    \betweenEdges{t-2}{t-1}
    \stepEdges{t-1}{t-2}{\theTimestep-1}{\theTimestep-2}
    \betweenEdges{t-1}{t}
    \stepEdges{t}{t-1}{\theTimestep}{\theTimestep-1}
    \betweenEdges{t}{t+1}
    \stepEdges{t+1}{t}{\theTimestep+1}{\theTimestep}
    \betweenEdges{t+1}{t+2}
\end{tikzpicture}}
    \caption{Conceptual diagram of the controller-stack interface for RNNs and LSTMs, unrolled across a portion of time and with $\theNumLayers = 2$ layers. The dashed lines indicate our proposed short-circuit connection from the stack reading to the output.}
    \label{fig:stack-rnn-diagram}
\end{figure}

Note that in this setup, when the hidden state issues actions at step $\theTimestep$ that update the stack, it cannot access the updated stack reading until step $\theTimestep+1$. This essentially causes an off-by-one lag between the model's predictions and stack state. Indeed, the proof of \citeposs{dusell-chiang-2023-surprising} Prop.~1, which shows that nondeterministic stack RNNs can recognize all context-free languages, relies on a final $\eos$ symbol to work around this. Although it may be possible for a model to learn a workaround, doing so likely overcomplicates simple parsing tasks and hinders hierarchical generalization. We propose the following fix. At step $\theTimestep$, we simply short-circuit $\theStackReadingT{\theTimestep}$ to the output $\theRNNOutputT{\theTimestep}$ by concatenating it to the final layer's hidden state. More precisely, $\theRNNOutputT{\theTimestep} \defeq \dropout{\begin{bmatrix}
    \theHiddenStateLT{\theNumLayers}{\theTimestep} \\
    \theStackReadingT{\theTimestep}
\end{bmatrix}}$. We also adjust the size of the embedding vectors in $\theInputEmbeddingMatrix$ to match the new size of $\theRNNOutputT{\theTimestep}$, so that $\theInputEmbeddingMatrix \in \realSet^{(|\theAlphabet|+1) \times (\theModelSize + \theStackReadingSize)}$, where $\theStackReadingSize$ is the size of $\theStackReadingT{\theTimestep}$.

\subsection{Stack Attention Transformer}

A transformer encoder consists of multiple layers, each of which consists of an attention sublayer followed by a feedforward sublayer. Since the differentiable stacks described earlier resemble structured attention mechanisms over syntactic structures, \citet{dusell-chiang-2024-stack} proposed incorporating them into the transformer by swapping them in place of the scaled dot-product attention operator in some of the attention sublayers. In one of these so-called \newterm{stack attention} sublayers, the differentiable stack starts in an empty configuration, and the sublayer inputs are used to compute actions that operate on the stack at each position in succession. The outputs of the stack attention layer are the intermediate stack readings at each position. See \cref{sec:stack-attention-details} and \citet{dusell-chiang-2024-stack} for details.

\begin{figure*}[t]
    \centering
    \begin{tikzpicture}
      \node[draw=green!60!black, fill=green!40, minimum width=10pt, minimum height=10pt] (hierarchical square) {};
      \node[right=0em of hierarchical square] (hierarchical text) {Hierarchical};
      \node[draw=red!60!black, fill=red!40, minimum width=10pt, minimum height=10pt, right=1em of hierarchical text] (linear square) {};
      \node[right=0em of linear square] (linear text) {Linear};
    \end{tikzpicture}
    \newcommand{\newrow}{\\[-12pt]}
    \begin{tabular}{cc}
        \begin{tikzpicture}
    \begin{axis}[
        mybarchart,
        ylabel={Full Accuracy},
        symbolic x coords={\transformerLabel{},\transformerLabel{}+\supLabel{},\transformerLabel{}+\supLabel{}+\supLabel{},\transformerLabel{}+\ndLabel{},\transformerLabel{}+\ndLabel{}+\ndLabel{}},
        title={Question Formation},
    ]
        \addplot+[hierarchical] coordinates {
(\transformerLabel{},0.005) +- (0,.01)
(\transformerLabel{}+\supLabel{},0.039) +- (0,.04)
(\transformerLabel{}+\supLabel{}+\supLabel{},0.161) +- (0,.18)
(\transformerLabel{}+\ndLabel{},0.318) +- (0,.19)
(\transformerLabel{}+\ndLabel{}+\ndLabel{},0.191) +- (0,.22)
        };
        \addplot+[linear] coordinates {
(\transformerLabel{},0.112) +- (0,.06)
(\transformerLabel{}+\supLabel{},0.202) +- (0,.15)
(\transformerLabel{}+\supLabel{}+\supLabel{},0.222) +- (0,.10)
(\transformerLabel{}+\ndLabel{},0.018) +- (0,.02)
(\transformerLabel{}+\ndLabel{}+\ndLabel{},0.067) +- (0,.13)
        };
    \end{axis}
\end{tikzpicture} & \begin{tikzpicture}
    \begin{axis}[
        mybarchart,
        ylabel={Partial Accuracy},
        symbolic x coords={\transformerLabel{},\transformerLabel{}+\supLabel{},\transformerLabel{}+\supLabel{}+\supLabel{},\transformerLabel{}+\ndLabel{},\transformerLabel{}+\ndLabel{}+\ndLabel{}},
        title={Question Formation},
    ]
        \addplot+[hierarchical] coordinates {
(\transformerLabel{},0.645) +- (0,.25)
(\transformerLabel{}+\supLabel{},0.325) +- (0,.21)
(\transformerLabel{}+\supLabel{}+\supLabel{},0.697) +- (0,.08)
(\transformerLabel{}+\ndLabel{},0.732) +- (0,.15)
(\transformerLabel{}+\ndLabel{}+\ndLabel{},0.862) +- (0,.17)
        };
        \addplot+[linear] coordinates {
(\transformerLabel{},0.355) +- (0,.25)
(\transformerLabel{}+\supLabel{},0.675) +- (0,.21)
(\transformerLabel{}+\supLabel{}+\supLabel{},0.303) +- (0,.08)
(\transformerLabel{}+\ndLabel{},0.268) +- (0,.15)
(\transformerLabel{}+\ndLabel{}+\ndLabel{},0.135) +- (0,.16)
        };
    \end{axis}
\end{tikzpicture} \newrow
        \begin{tikzpicture}
    \begin{axis}[
        mybarchart,
        ylabel={Full Accuracy},
        symbolic x coords={\transformerLabel{},\transformerLabel{}+\supLabel{},\transformerLabel{}+\supLabel{}+\supLabel{},\transformerLabel{}+\ndLabel{},\transformerLabel{}+\ndLabel{}+\ndLabel{}},
        title={German Question Formation},
    ]
        \addplot+[hierarchical] coordinates {
(\transformerLabel{},0.000) +- (0,.00)
(\transformerLabel{}+\supLabel{},0.000) +- (0,.00)
(\transformerLabel{}+\supLabel{}+\supLabel{},0.000) +- (0,.00)
(\transformerLabel{}+\ndLabel{},0.004) +- (0,.01)
(\transformerLabel{}+\ndLabel{}+\ndLabel{},0.001) +- (0,.00)
        };
        \addplot+[linear] coordinates {
(\transformerLabel{},0.000) +- (0,.00)
(\transformerLabel{}+\supLabel{},0.000) +- (0,.00)
(\transformerLabel{}+\supLabel{}+\supLabel{},0.000) +- (0,.00)
(\transformerLabel{}+\ndLabel{},0.000) +- (0,.00)
(\transformerLabel{}+\ndLabel{}+\ndLabel{},0.000) +- (0,.00)
        };
    \end{axis}
\end{tikzpicture} & \begin{tikzpicture}
    \begin{axis}[
        mybarchart,
        ylabel={Partial Accuracy},
        symbolic x coords={\transformerLabel{},\transformerLabel{}+\supLabel{},\transformerLabel{}+\supLabel{}+\supLabel{},\transformerLabel{}+\ndLabel{},\transformerLabel{}+\ndLabel{}+\ndLabel{}},
        title={German Question Formation},
    ]
        \addplot+[hierarchical] coordinates {
(\transformerLabel{},0.868) +- (0,.26)
(\transformerLabel{}+\supLabel{},0.999) +- (0,.00)
(\transformerLabel{}+\supLabel{}+\supLabel{},0.996) +- (0,.01)
(\transformerLabel{}+\ndLabel{},1.000) +- (0,.00)
(\transformerLabel{}+\ndLabel{}+\ndLabel{},0.979) +- (0,.04)
        };
        \addplot+[linear] coordinates {
(\transformerLabel{},0.003) +- (0,.01)
(\transformerLabel{}+\supLabel{},0.001) +- (0,.00)
(\transformerLabel{}+\supLabel{}+\supLabel{},0.004) +- (0,.01)
(\transformerLabel{}+\ndLabel{},0.000) +- (0,.00)
(\transformerLabel{}+\ndLabel{}+\ndLabel{},0.000) +- (0,.00)
        };
    \end{axis}
\end{tikzpicture} \newrow
        \begin{tikzpicture}
    \begin{axis}[
        mybarchart,
        ylabel={Full Accuracy},
        symbolic x coords={\transformerLabel{},\transformerLabel{}+\supLabel{},\transformerLabel{}+\supLabel{}+\supLabel{},\transformerLabel{}+\ndLabel{},\transformerLabel{}+\ndLabel{}+\ndLabel{}},
        title={Tense Reinflection with \textit{Do}},
    ]
        \addplot+[hierarchical] coordinates {
(\transformerLabel{},0.145) +- (0,.15)
(\transformerLabel{}+\supLabel{},0.368) +- (0,.10)
(\transformerLabel{}+\supLabel{}+\supLabel{},0.127) +- (0,.12)
(\transformerLabel{}+\ndLabel{},0.416) +- (0,.10)
(\transformerLabel{}+\ndLabel{}+\ndLabel{},0.165) +- (0,.11)
        };
        \addplot+[linear] coordinates {
(\transformerLabel{},0.852) +- (0,.14)
(\transformerLabel{}+\supLabel{},0.629) +- (0,.10)
(\transformerLabel{}+\supLabel{}+\supLabel{},0.871) +- (0,.12)
(\transformerLabel{}+\ndLabel{},0.521) +- (0,.14)
(\transformerLabel{}+\ndLabel{}+\ndLabel{},0.829) +- (0,.12)
        };
    \end{axis}
\end{tikzpicture} & \begin{tikzpicture}
    \begin{axis}[
        mybarchart,
        ylabel={Partial Accuracy},
        symbolic x coords={\transformerLabel{},\transformerLabel{}+\supLabel{},\transformerLabel{}+\supLabel{}+\supLabel{},\transformerLabel{}+\ndLabel{},\transformerLabel{}+\ndLabel{}+\ndLabel{}},
        title={Tense Reinflection with \textit{Do}},
    ]
        \addplot+[hierarchical] coordinates {
(\transformerLabel{},0.145) +- (0,.15)
(\transformerLabel{}+\supLabel{},0.370) +- (0,.10)
(\transformerLabel{}+\supLabel{}+\supLabel{},0.128) +- (0,.12)
(\transformerLabel{}+\ndLabel{},0.476) +- (0,.15)
(\transformerLabel{}+\ndLabel{}+\ndLabel{},0.167) +- (0,.12)
        };
        \addplot+[linear] coordinates {
(\transformerLabel{},0.854) +- (0,.15)
(\transformerLabel{}+\supLabel{},0.630) +- (0,.10)
(\transformerLabel{}+\supLabel{}+\supLabel{},0.871) +- (0,.12)
(\transformerLabel{}+\ndLabel{},0.523) +- (0,.15)
(\transformerLabel{}+\ndLabel{}+\ndLabel{},0.831) +- (0,.12)
        };
    \end{axis}
\end{tikzpicture} \newrow
        \begin{tikzpicture}
    \begin{axis}[
        mybarchart,
        ylabel={Full Accuracy},
        symbolic x coords={\transformerLabel{},\transformerLabel{}+\supLabel{},\transformerLabel{}+\supLabel{}+\supLabel{},\transformerLabel{}+\ndLabel{},\transformerLabel{}+\ndLabel{}+\ndLabel{},\rnnLabel{}},
        title={Passivization},
    ]
        \addplot+[hierarchical] coordinates {
(\transformerLabel{},0.000) +- (0,.00)
(\transformerLabel{}+\supLabel{},0.000) +- (0,.00)
(\transformerLabel{}+\supLabel{}+\supLabel{},0.000) +- (0,.00)
(\transformerLabel{}+\ndLabel{},0.000) +- (0,.00)
(\transformerLabel{}+\ndLabel{}+\ndLabel{},0.000) +- (0,.00)
(\rnnLabel{},0.000) +- (0,.00)
        };
        \addplot+[linear] coordinates {
(\transformerLabel{},0.000) +- (0,.00)
(\transformerLabel{}+\supLabel{},0.000) +- (0,.00)
(\transformerLabel{}+\supLabel{}+\supLabel{},0.001) +- (0,.00)
(\transformerLabel{}+\ndLabel{},0.000) +- (0,.00)
(\transformerLabel{}+\ndLabel{}+\ndLabel{},0.000) +- (0,.00)
(\rnnLabel{},0.000) +- (0,.00)
        };
    \end{axis}
\end{tikzpicture} & \begin{tikzpicture}
    \begin{axis}[
        mybarchart,
        ylabel={Partial Accuracy},
        symbolic x coords={\transformerLabel{},\transformerLabel{}+\supLabel{},\transformerLabel{}+\supLabel{}+\supLabel{},\transformerLabel{}+\ndLabel{},\transformerLabel{}+\ndLabel{}+\ndLabel{},\rnnLabel{}},
        title={Passivization},
    ]
        \addplot+[hierarchical] coordinates {
(\transformerLabel{},0.526) +- (0,.34)
(\transformerLabel{}+\supLabel{},0.565) +- (0,.31)
(\transformerLabel{}+\supLabel{}+\supLabel{},0.758) +- (0,.37)
(\transformerLabel{}+\ndLabel{},0.612) +- (0,.46)
(\transformerLabel{}+\ndLabel{}+\ndLabel{},0.569) +- (0,.38)
(\rnnLabel{},0.061) +- (0,.01)
        };
        \addplot+[linear] coordinates {
(\transformerLabel{},0.196) +- (0,.18)
(\transformerLabel{}+\supLabel{},0.144) +- (0,.19)
(\transformerLabel{}+\supLabel{}+\supLabel{},0.276) +- (0,.37)
(\transformerLabel{}+\ndLabel{},0.383) +- (0,.42)
(\transformerLabel{}+\ndLabel{}+\ndLabel{},0.280) +- (0,.36)
(\rnnLabel{},0.146) +- (0,.02)
        };
    \end{axis}
\end{tikzpicture}
    \end{tabular}
    \caption{Full and partial accuracy on the hierarchical and linear generalization sets of models that achieved at least 90\% accuracy on the test set. All bars show the mean of 5 runs; error bars indicate one standard deviation. ``\transformerLabel{}'' = transformer; ``\rnnLabel{}'' = simple RNN; ``+\supLabel{}'' = with superposition stack; ``+\ndLabel{}'' = with nondeterministic stack.}
    \label{fig:bar-charts}
\end{figure*}

\section{Experiments}
\label{sec:experiments}

We train stack-augmented and vanilla neural networks to transform an input string $\theInputString$ to an output string $\theOutputString$ for the four tasks introduced in \cref{sec:tasks}. Following \citet{murty-etal-2023-grokking} and \citet{ahuja-etal-2025-learning}, we train all models as autoregressive language models on the concatenation of the input and output, $\theString = \theInputString \theOutputString$.
Prompting a model $\theModel$ with a prefix $\theInputString$ allows us to sample outputs from or measure the probabilities of outputs in the conditional distribution $\theModelProbDist(\cdot \mid \theInputString)$.

We automatically adjust $\theModelSize$ for each model so that its total parameter count is as close as possible to 200k, ensuring that all models are of comparable size. We use SymPy \citep{meurer-etal-2017-sympy} to work out the algebra automatically. All RNNs and LSTMs have 3 layers, and all transformers have 5 layers. We use transformers with either one or two stack attention layers. For transformers with one stack attention layer, we swap it into the attention mechanism of the third layer. For transformers with two stack attention layers, we swap them into the second and fourth layers. All superposition stacks use a stack vector size of $\theStackVectorSize = 50$. All nondeterministic stacks use $|\theStateSet| = 3$, $|\theStackAlphabet| = 3$, and $\theStackVectorSize = 5$.

We have included transformers with two stack attention layers for the following reason. Even with a stack, an autoregressive language model is still somewhat at a disadvantage on these tasks, in that the model not only needs to parse the input, but also to continue generating a transformed version of it. This is akin to modeling cross-serial dependencies, for which a stack is inadequate; if the language of inputs admitted arbitrary levels of recursion, then the language of concatenated inputs and outputs would not even be context-free. \Citet{dusell-chiang-2023-surprising} showed that nondeterministic stack LSTMs can actually learn cross-serial dependencies, but stack transformers cannot perform the same trick with a single layer. Thus, it might be helpful for the transformer to have a stack attention layer that parses the input, an intermediate standard attention layer that copies it to the output, and another stack attention layer that models the syntax of the output.

We optimize each model using a standard cross-entropy objective. For each architecture, we do an initial hyperparameter search by training 10 models with randomly sampled hyperparameters for a maximum of 5 epochs each. We select the hyperparameters that result in the lowest cross-entropy on the validation set and then train 5 models to convergence. We report the mean and standard deviation of the performance of these models. See \cref{sec:experimental-details} for details.

We also make some improvements upon previously used metrics. Instead of evaluating the accuracy of greedily decoded outputs, which can obscure the learned distribution, we measure \emph{expected} accuracy with respect to $\theModelProbDist(\cdot \mid \theInputString)$. This gives us the probability of generating a particular output using ancestral sampling, which better matches how modern language models are used. We measure both \newterm{full accuracy (FA)} (checking whether the entire output matches) and \newterm{partial accuracy (PA)} (checking whether some salient, task-specific aspect of the output matches, if not the whole thing). We report accuracy with respect to the outputs predicted by both the hierarchical and the linear rules, allowing us to compare
the two (since they are only two of many possible rules, their accuracies can sum to less than 1).

The expected full accuracy with respect to output $\theOutputString$ is simply the probability $\theModelProbDist(\theOutputString \mid \theInputString)$, which we can compute exactly. Because the transformation is deterministic conditioned on the input, a probability of 1 is achievable.
Let $\theDataset$ be a dataset of input-output pairs, which can be the in-distribution test set, the generalization set, or the generalization set according to the incorrect linear rule.
\begin{equation}
    \fullAccuracy{\theModel}{\theDataset} \defeq \frac{1}{|\theDataset|} \sum_{(\theInputString, \theOutputString) \in \theDataset} \theConditionalProbability
\end{equation}

We use the same task-specific partial accuracy metrics as prior work but compute them in expectation. For question formation (both English and German), we use first-word accuracy, which tests whether the model fronted the correct verb. We compute the expectation exactly, by looking at the conditional distribution over words after reading $\theInputString$ and averaging the probability of the expected first word. For tense reinflection, we test whether the output is of the correct length, has words that are all of the correct parts of speech in the correct positions (even if the exact word choices do not match), and that the surface form of the reinflected main verb is correct.
We estimate the expectation by randomly sampling 10 outputs for each input and computing the proportion of outputs that partially match. For passivization, we use second-word accuracy, which tests whether the model moved the correct noun to the subject position. Again, we estimate the expectation by randomly sampling 10 outputs for each input. Note that the generalization set for passivization contains examples where the second noun and the object noun happen to be the same word type. Therefore, it is possible for the partial accuracy with respect to the hierarchical and linear rules to sum to more than 1.

\section{Results}
\label{sec:results}

On all tasks, only transformers learn the in-distribution test set.
The RNNs and LSTMs likely fail to learn the test set because their hidden states, which are of constant size, impose an information bottleneck between the input and output, hindering the models from copying parts of the input to the output faithfully. The transformers do not have this bottleneck thanks to their self-attention mechanism.\footnote{Unlike our RNN and LSTM language models, which do not have self-attention, the RNN, GRU, and LSTM models of \citet{mccoy-etal-2020-syntax} were encoder-decoder models with cross-attention, hence they were sometimes able to learn the test set.} Since we are primarily interested in models that succeed both in-distribution and out-of-distribution, in \cref{fig:bar-charts}, we report results only for models with at least 90\% test accuracy. We provide results for all models and tasks in \cref{sec:full-results}.

The vanilla transformer (\transformerLabel{}) has linear FA but hierarchical PA on question formation. All transformers, even \transformerLabel{}, prefer the hierarchical rule on German question formation and passivization, but they prefer the linear rule on tense reinflection with \textit{do}. The transformer with nondeterministic stack attention (\transformerLabel{}+\ndLabel{}) attains strong performance on all four tasks. On question formation, it is the best model and, unlike \transformerLabel{}, prefers to generate \emph{full} outputs according to the hierarchical rule. Unlike typical results in prior work that generate the correct full output 0\% of the time, \transformerLabel{}+\ndLabel{} does so an average of 32\% of the time. This result is remarkable because \citet{mccoy-etal-2020-syntax} had to go to considerable lengths to achieve even partial-accuracy hierarchical generalization on the same task; they used a Tree-GRU encoder-decoder whose computation graph was structured according to the correct parse tree, not only on the input side, but also on the \emph{output} side. Merely marking the correct parse tree structure using brackets was not enough for standard architectures to generalize hierarchically. Even the ON-LSTM \citep{shen-etal-2018-ordered}, a syntactically unsupervised architecture that simulates a stack of bounded depth, failed to generalize.

\transformerLabel{}+\ndLabel{} also has perfect PA on German question formation, and it is the best model on tense reinflection with \textit{do} (although it still prefers the linear rule). It is also better than \transformerLabel{} on passivization, although it has lower absolute PA than \transformerLabel{}+\supLabel{}+\supLabel{}. Adding another nondeterministic stack attention layer (\transformerLabel{}+\ndLabel{}+\ndLabel{}) sometimes helps and sometimes hurts. It has even better PA on question formation than \transformerLabel{}+\ndLabel{}, also beating the GRU encoder-decoder with location-based attention reported by \citet{mccoy-etal-2020-syntax}, the ON-LSTM reported by \citet{mccoy-etal-2020-syntax}, and the overtrained transformers reported by \citet{ahuja-etal-2025-learning} and \citet{qin-etal-2025-data}, but not \citet{murty-etal-2023-grokking}. The \transformerLabel{}+\supLabel{} model is better than \transformerLabel{} on all tasks except question formation. Adding another superposition stack attention layer (\transformerLabel{}+\supLabel{}+\supLabel{}) has mixed results, proving beneficial only for question formation and passivization. In general, nondeterministic stacks have an edge over superposition stacks. On German question formation, all stack transformers outperform \transformerLabel{}, the pre-trained model mBART reported by \citet{mueller-etal-2022-coloring}, and the overtrained transformer reported by \citet{ahuja-etal-2025-learning}. The results for adding shortcut connections to the RNNs and LSTMs are mixed; for German question formation (\cref{tab:question-formation-in-german}), it always improves hierarchical generalization except for \rnnLabel{}+\ndLabel{} PA, and for other tasks it does not consistently help.

\section{Conclusion}

Our results show that on multiple tasks, especially question formation, transformers with nondeterministic stack attention show a much stronger hierarchical bias than standard architectures, putting them more in line with human inductive biases. \Citet{dusell-chiang-2024-stack} showed that the same architecture outperforms the vanilla transformer on a Penn Treebank language modeling task despite having fewer parameters, so the nondeterministic stack attention transformer not only generalizes more naturally in terms of syntax, but fits in-distribution natural language better, suggesting that it is a promising object of future psycholinguistic study.
Stack-augmented neural networks do not appear to be a silver bullet for hierarchical generalization; like any learning algorithm, they are likely sensitive to properties of the data (cf.~\citet{qin-etal-2025-data}), as seen by the variation of results across tasks. However, on the whole they have a higher tendency than standard architectures to generalize hierarchically without explicit hints.

\section*{Limitations}

Our experiments, which use previously established datasets, include only two languages: English and German. More comprehensive experiments would include syntactic phenomena from more languages from diverse language families.

\ifAnonymousSubmissionMode
\else
\section*{Acknowledgments}

We thank Tal Linzen for helpful feedback on our experiments. We thank Alexander Hoyle, Tianyu Liu, Irene Strauss, Alexandra Butoi, Karolina Sta{\'n}czak, and Blanka K{\"o}v{\'e}r for comments on earlier drafts of this paper.
\fi

\bibliography{bibliography}

\appendix

\section{Issues with German Passivization}
\label{sec:passivization-in-german-issues}

We have omitted the German passivization task of \citet{mueller-etal-2022-coloring} from our main results because of issues in the task design; we have also found and corrected bugs in the generalization set.

\paragraph{Task Design Issues.} In the training data, the second noun phrase is always a direct object in accusative case. However, in the generalization data, the second noun phrase is the object of a preposition in \emph{dative case}. Therefore, the surface form of the second noun phrase does not even match the supposedly ambiguous training data rule. This may lead to spuriously high hierarchical generalization accuracy. As expected, in \cref{tab:passivization-in-german}, the baseline transformer already has a strong preference for the hierarchical rule. Furthermore, in some cases, the number of the direct object is inherently ambiguous (e.g., \textit{den L{\"o}wen} can be either accusative singular or dative plural of \textit{der L{\"o}we}), so two different passive outputs are possible. It seems that the task assumes accusative singular in ambiguous cases and so inflects it as nominative singular when moving it to subject position; but this transformation also has a bug, as described below.

\paragraph{Bugs in the Generalization Set.} In transformed outputs, \textit{der L{\"o}we} and \textit{der Rabe} are declined incorrectly as \textit{der L{\"o}wen} and \textit{der Raben} in the nominative singular. Our results in \cref{tab:passivization-in-german} use a version of the dataset in which we have fixed this bug.

\section{Details of Differentiable Stacks}

\subsection{Superposition Stack}
\label{sec:superposition-stack-details}

Let $\theStackActionVectorT{1}, \ldots, \theStackActionVectorT{\theTimestep}$ be a sequence of stack action vectors, and let $\thePushedVectorT{1}, \ldots, \thePushedVectorT{\theTimestep}$ be the corresponding pushed vectors. Let a \newterm{run} $\theRun = \theStackActionT{\theOtherTimestep}, \ldots, \theStackActionT{\theOtherOtherTimestep}$, where $\theStackActionT{\theOtherTimestep}, \ldots, \theStackActionT{\theOtherOtherTimestep} \in \{ \pushLabel, \noopLabel, \popLabel \}$, be any sequence of actions that, if executed on an empty stack, would never attempt to pop the stack when it was empty. Let $\weightOfRun{\theRun}$ be the weight of $\theRun$, which is the product $\theSupActionProbT{\theOtherTimestep}{\theStackActionT{\theOtherTimestep}} \cdots \theSupActionProbT{\theOtherOtherTimestep}{\theStackActionT{\theOtherOtherTimestep}}$. Let $\topVectorOfRun{\theRun}$ be the top stack element that would result from starting with a stack containing just $\zeroVector$ and executing the actions in $\theRun$ on it, using $\thePushedVectorT{\theTimestep}$ as the pushed vector whenever $\theStackActionT{\theTimestep} = \pushLabel$. Let $\supRunSet{\theTimestep}$ be the set of all runs that start with a $\pushLabel$ and end at step $\theTimestep$. Then the stack reading can be expressed as
\begin{equation}
    \label{eq:sup-stack-reading}
    \theStackReadingT{\theTimestep} = \frac{\sum_{\theRun \in \supRunSet{\theTimestep}} \weightOfRun{\theRun} \, \topVectorOfRun{\theRun}}{\sum_{\theRun \in \supRunSet{\theTimestep}} \weightOfRun{\theRun}}.
\end{equation}
The procedure described in \cref{sec:superposition-stack} is an efficient dynamic programming algorithm for computing \cref{eq:sup-stack-reading}. \Citet{dusell-chiang-2024-stack} pointed out that, mathematically, the stack reading at step $\theTimestep$ is simply a summation over all possible sequences of stack actions ending at $\theTimestep$, so it resembles a kind of structured attention mechanism over $\thePushedVectorT{1}, \ldots, \thePushedVectorT{\theTimestep}$ based on syntactic structures.

\subsection{Nondeterministic Stack}
\label{sec:nondeterministic-stack-details}

In the dVPDA, each transition $\theTransition$ has a non-negative weight $\weightOfTransition{\theTransition}$; accordingly, the action vector $\theStackActionVectorT{\theTimestep} = \exp(\theStackActionLogitsT{\theTimestep})$ contains the weights of all transitions corresponding to $\theSymbolT{\theTimestep}$, of which there are a fixed number. Let a \newterm{run} $\theRun = \theTransitionT{1}, \ldots, \theTransitionT{\theOtherOtherTimestep}$ be any valid sequence of transitions that starts in the initial configuration and never fully empties the stack (replacing the bottom element does not count). Let $\weightOfRun{\theRun}$ be the weight of $\theRun$, which is the product $\weightOfTransition{\theTransitionT{1}} \cdots \weightOfTransition{\theTransitionT{\theOtherOtherTimestep}}$. Let $\topVectorOfRun{\theRun}$ be the top stack vector that would result at the end of $\theRun$. Let $\ndRunSet{\theTimestep}{\theFinalState}{\theTopStackSymbol}$ be the set of all runs that end at step $\theTimestep$ in state $\theFinalState$ and with $\theTopStackSymbol$ on top of the stack. For each $\theFinalState, \theTopStackSymbol, \theTimestep$, the dVPDA computes a summation over all of those runs.
\begin{equation}
    \label{eq:nd-stack-reading}
    \begin{split}
        &\theNDStackReadingSlice{\theTimestep}{\theFinalState}{\theTopStackSymbol} \defeq \\
        &\quad \frac{
            \sum_{\theRun \in \ndRunSet{\theTimestep}{\theFinalState}{\theTopStackSymbol}} \weightOfRun{\theRun} \, \topVectorOfRun{\theRun}
        }{
            \sum_{\theFinalState' \in \theStateSet} \sum_{\theTopStackSymbol' \in \theStackAlphabet} \sum_{\theRun \in \ndRunSet{\theTimestep}{\theFinalState'}{\theTopStackSymbol'}} \weightOfRun{\theRun}
        }
    \end{split}
\end{equation}
Again, this resembles a structured attention mechanism over syntactic structures. \Citet{dusell-chiang-2023-surprising} show how to compute \cref{eq:nd-stack-reading} efficiently using a dynamic programming algorithm. The stack reading $\theStackReadingT{\theTimestep}$ is the concatenation of all $\theNDStackReadingSlice{\theTimestep}{\theFinalState}{\theTopStackSymbol}$. Note that \cref{eq:sup-stack-reading} is a special case of \cref{eq:nd-stack-reading} where $\theStateSet = \{\theInitialState\}$, $\theStackAlphabet = \{\stackBottomSymbol\}$, $\thePushedVectorT{0} = \zeroVector$, transition weights are locally normalized to sum to 1, and replace transitions before the first push in a run have a weight of 1.

\section{Details of Neural Network Architectures}
\label{sec:base-nn-details}

In this section, we describe each of the base neural network architectures (simple RNN, LSTM, and transformer) in more detail. All three are based on the built-in PyTorch implementations \citep{paszke-etal-2019-pytorch} and closely follow the settings of \citet{butoi-etal-2025-training}. Each architecture consists of a configurable number of layers $\theNumLayers$. Each architecture uses a learned input embedding matrix $\theInputEmbeddingMatrix$ to map each input symbol $\theSymbolT{\theTimestep}$ of the input string to an embedding $\theInputEmbeddingT{\theTimestep} = \theInputEmbeddingMatrix_{\theSymbolT{\theTimestep}}$. The size of the embeddings is $\theModelSize$.

\subsection{Simple RNN}
\label{sec:rnn-details}

Let $\theHiddenStateLT{\theLayerNo}{\theTimestep}$ denote the hidden state of the $\theLayerNo$\textsuperscript{th} layer at timestep $\theTimestep$. We apply dropout to the input embeddings, the hidden states between layers, and the hidden states of the last layer, following \citet{zaremba-etal-2015-recurrent}. The initial hidden state of each layer is learned. The simple RNN architecture is defined in \cref{fig:simple-rnn-equations}. The learned parameters are $\theInitialHiddenStateParamL{\theLayerNo} \in \realSet^{\theModelSize}$, $\weightParamL{h}{\theLayerNo} \in \realSet^{\theModelSize \times 2\theModelSize}$, and $\biasParamL{h}{\theLayerNo} \in \realSet^{\theModelSize}$ ($1 \leq \theLayerNo \leq \theNumLayers$).

\begin{figure*}[t]
    \centering
    \begin{subequations}
    \begin{align}
        \theRNNInputT{\theTimestep} &\defeq \dropout{\theInputEmbeddingT{\theTimestep}} && (1 \leq \theTimestep \leq \theLength) \\
        \theHiddenStateDropoutLT{0}{\theTimestep} &\defeq \theRNNInputT{\theTimestep} && (1 \leq \theTimestep \leq \theLength) \\
        \theHiddenStateLT{\theLayerNo}{0} &\defeq \tanh(\theInitialHiddenStateParamL{\theLayerNo}) \quad && (1 \leq \theLayerNo \leq \theNumLayers) \\
        \theHiddenStateLT{\theLayerNo}{\theTimestep} &\defeq \tanh(\affineL{h}{\theLayerNo}{\begin{bmatrix} \theHiddenStateDropoutLT{\theLayerNo-1}{\theTimestep} \\ \theHiddenStateLT{\theLayerNo}{\theTimestep-1} \end{bmatrix}}) \quad && (1 \leq \theLayerNo \leq \theNumLayers; 1 \leq \theTimestep \leq \theLength) \\
        \theHiddenStateDropoutLT{\theLayerNo}{\theTimestep} &\defeq \dropout{\theHiddenStateLT{\theLayerNo}{\theTimestep}} && (1 \leq \theLayerNo \leq \theNumLayers - 1; 1 \leq \theTimestep \leq \theLength) \\
        \theRNNStateT{\theTimestep} &\defeq (\theHiddenStateLT{1}{\theTimestep}, \ldots, \theHiddenStateLT{\theNumLayers}{\theTimestep}) && (0 \leq \theTimestep \leq \theLength) \\
        \theRNNOutputT{\theTimestep} &\defeq \dropout{\theHiddenStateLT{\theNumLayers}{\theTimestep}} && (0 \leq \theTimestep \leq \theLength)
    \end{align}
    \end{subequations}
    \caption{Equations for the simple RNN.}
    \label{fig:simple-rnn-equations}
\end{figure*}

\subsection{LSTM}
\label{sec:lstm-details}

We apply dropout according to \citet{zaremba-etal-2015-recurrent}. The initial hidden state of each layer is learned. The LSTM architecture is defined as follows. Let $\elementwiseMultiply$ denote elementwise multiplication. The LSTM architecture is defined in \cref{fig:lstm-equations}. The learned parameters are $\theInitialHiddenStateParamL{\theLayerNo} \in \realSet^{\theModelSize}$, $\weightParamL{i}{\theLayerNo} \in \realSet^{\theModelSize \times 2\theModelSize}$, $\biasParamL{i}{\theLayerNo} \in \realSet^{\theModelSize}$, $\weightParamL{f}{\theLayerNo} \in \realSet^{\theModelSize \times 2\theModelSize}$, $\biasParamL{f}{\theLayerNo} \in \realSet^{\theModelSize}$, $\weightParamL{g}{\theLayerNo} \in \realSet^{\theModelSize \times 2\theModelSize}$, $\biasParamL{g}{\theLayerNo} \in \realSet^{\theModelSize}$, $\weightParamL{o}{\theLayerNo} \in \realSet^{\theModelSize \times 2\theModelSize}$, and $\biasParamL{o}{\theLayerNo} \in \realSet^{\theModelSize}$ ($1 \leq \theLayerNo \leq \theNumLayers$).

\begin{figure*}[t]
    \centering
    \newcommand{\lstmaffine}[1]{\affineL{#1}{\theLayerNo}{\begin{bmatrix} \theHiddenStateDropoutLT{\theLayerNo-1}{\theTimestep} \\ \theHiddenStateLT{\theLayerNo}{\theTimestep-1} \end{bmatrix}}}
    \newcommand{\bounds}{(1 \leq \theLayerNo \leq \theNumLayers; 1 \leq \theTimestep \leq \theLength)}
    \begin{subequations}
    \begin{align}
        \theRNNInputT{\theTimestep} &\defeq \dropout{\theInputEmbeddingT{\theTimestep}} && (1 \leq \theTimestep \leq \theLength) \\
        \theHiddenStateDropoutLT{0}{\theTimestep} &\defeq \theRNNInputT{\theTimestep} && (1 \leq \theTimestep \leq \theLength) \\
        \theHiddenStateLT{\theLayerNo}{0} &\defeq \tanh(\theInitialHiddenStateParamL{\theLayerNo}) \quad && (1 \leq \theLayerNo \leq \theNumLayers) \\
        \theMemoryCellLT{\theLayerNo}{0} &\defeq \zeroVector \quad && (1 \leq \theLayerNo \leq \theNumLayers) \\
        \theInputGateLT{\theLayerNo}{\theTimestep} &\defeq \logistic{\lstmaffine{i}} \quad && \bounds \\
        \theForgetGateLT{\theLayerNo}{\theTimestep} &\defeq \logistic{\lstmaffine{f}} \quad && \bounds \\
        \theCandidateLT{\theLayerNo}{\theTimestep} &\defeq \tanh(\lstmaffine{g}) \quad && \bounds \\
        \theOutputGateLT{\theLayerNo}{\theTimestep} &\defeq \logistic{\lstmaffine{o}} \quad && \bounds \\
        \theMemoryCellLT{\theLayerNo}{\theTimestep} &\defeq \theForgetGateLT{\theLayerNo}{\theTimestep} \elementwiseMultiply \theMemoryCellLT{\theLayerNo}{\theTimestep-1} + \theInputGateLT{\theLayerNo}{\theTimestep} \elementwiseMultiply \theCandidateLT{\theLayerNo}{\theTimestep} \quad && \bounds \\
        \theHiddenStateLT{\theLayerNo}{\theTimestep} &\defeq \theOutputGateLT{\theLayerNo}{\theTimestep} \elementwiseMultiply \tanh(\theMemoryCellLT{\theLayerNo}{\theTimestep}) \quad && \bounds \\
        \theHiddenStateDropoutLT{\theLayerNo}{\theTimestep} &\defeq \dropout{\theHiddenStateLT{\theLayerNo}{\theTimestep}} && (1 \leq \theLayerNo \leq \theNumLayers - 1; 1 \leq \theTimestep \leq \theLength) \\
        \theRNNStateT{\theTimestep} &\defeq (\theHiddenStateLT{1}{\theTimestep}, \theMemoryCellLT{1}{\theTimestep} \ldots, \theHiddenStateLT{\theNumLayers}{\theTimestep}, \theMemoryCellLT{\theNumLayers}{\theTimestep}) && (0 \leq \theTimestep \leq \theLength) \\
        \theRNNOutputT{\theTimestep} &\defeq \dropout{\theHiddenStateLT{\theNumLayers}{\theTimestep}} && (0 \leq \theTimestep \leq \theLength)
    \end{align}
    \end{subequations}
    \caption{Equations for the LSTM.}
    \label{fig:lstm-equations}
\end{figure*}

\subsection{Transformer}
\label{sec:transformer-details}

Our transformer implementation is based on that of PyTorch. Following \citet{vaswani-etal-2017-attention}, we map input symbols to vectors of size $\theModelSize$ with a scaled embedding layer and add sinusoidal positional encodings. We always prepend a reserved $\bos$ token whose corresponding output is used as the probabilities for the first token. We tie the input and output embeddings \citep{press-wolf-2017-using}. We set the number of hidden units in each feedforward layer to $2 \theModelSize$. We use pre-norm instead of post-norm and apply layer norm to the output of the last layer. We use the same dropout rate throughout the transformer. We apply it in the same places as \citet{vaswani-etal-2017-attention}, and, as implemented by PyTorch, we also apply it to the hidden units of feedforward sublayers and to the attention probabilities of scaled dot-product attention operations.

\subsection{Stack Attention Transformer}
\label{sec:stack-attention-details}

Let $\theTfSublayerInputT{\theTimestep}$ be the input to a sublayer. In a transformer with pre-norm, we pass the input through layer norm \citep{ba-etal-2016-layer} to get $\theTfSublayerLNOutputT{\theTimestep} \defeq \layerNorm{\theTfSublayerInputT{\theTimestep}}$. Let $\sublayerFunctionName$ be the \newterm{sublayer function}, which encapsulates the core behavior of the sublayer, such as attention or a feedforward layer. Then $\theTfSublayerFuncOutputT{\theTimestep} \defeq \sublayerFunctionName(\theTimestep, (\theTfSublayerLNOutputT{1}, \ldots, \theTfSublayerLNOutputT{\theLength}))$, and the output is computed by applying dropout to the sublayer output and adding a residual connection, giving $\theTfSublayerOutputT{\theTimestep} \defeq \theTfSublayerInputT{\theTimestep} + \dropout{\theTfSublayerFuncOutputT{\theTimestep}}$.

A stack attention layer implements $\sublayerFunctionName$ as follows. We compute the stack action logits as $\theStackActionLogitsT{\theTimestep} \defeq \linearTransform{a}{\theTfSublayerLNOutputT{\theTimestep}}$ and the pushed vector as $\thePushedVectorT{\theTimestep} \defeq \logistic{\linearTransform{v}{\theTfSublayerLNOutputT{\theTimestep}}}$, where $\weightParam{a} \in \realSet^{\theNumActions \times \theModelSize}$ and $\weightParam{v} \in \realSet^{\theStackVectorSize \times \theModelSize}$ are learned parameters. The sublayer starts with an empty stack and applies the stack operations on it for $\theTimestep = 1, \ldots, \theLength$. The stack returns a stack reading $\theStackReadingT{\theTimestep}$ at each step, and we compute the sublayer output as $\theTfSublayerOutputT{\theTimestep} \defeq \linearTransform{y}{\theStackReadingT{\theTimestep}}$, where $\weightParam{y} \in \realSet^{\theModelSize \times \theStackReadingSize}$ is a learned parameter.
We illustrate this in \cref{fig:stack-attention-diagram}.

\begin{figure}[t]
    \centering
    \scalebox{0.8}{\begin{tikzpicture}[x=1cm,y=1cm,>=Stealth,baseline={(current bounding box.center)}]
    \tikzset{linear/.style={dotted}}
    \tikzset{residual/.style={dashed}}
    \newcommand{\dropoutLabel}{\scalebox{0.8}{$\funcName{Dropout}$}}
    \newcommand{\layerNormLabel}{\scalebox{0.8}{$\funcName{LayerNorm}$}}
    \def\xsep{2.2}
    \def\ysep{1.8}
    \def\ysepmid{0.9}
    \def\dotsx{3.5}
    \def\labelx{3.7}
    \newcommand{\mycolumn}[3]{%
        \node (h #2) at (#1*\xsep, -\ysep) {$\theTfSublayerInputT{#3}$};
        \node (ln #2) at (#1*\xsep, -\ysepmid) {\layerNormLabel};
        \draw[->] (h #2) edge (ln #2);
        \node (s #2) at (#1*\xsep, 0) {$\theStackObjectT{#3}$};
        \draw[->, linear] (ln #2) edge node[right, align=left] {$\theStackActionVectorT{#3}, \thePushedVectorT{#3}$} (s #2);
        \node (dr #2) at (#1*\xsep, \ysepmid) {\dropoutLabel};
        \draw[->, linear] (s #2) edge node[right] {$\theStackReadingT{#3}$} (dr #2);
        \node (o #2) at (#1*\xsep, \ysep) {$\theTfSublayerOutputT{#3}$};
        \draw[->] (dr #2) edge (o #2);
        \draw[->, bend left, residual] (h #2) edge (o #2);
    }

    \node (leftdots) at (-\dotsx, 0) {$\cdots$};
    \mycolumn{-1}{t-1}{\theTimestep-1}
    \mycolumn{0}{t}{\theTimestep}
    \mycolumn{1}{t+1}{\theTimestep+1}
    \node (rightdots) at (\dotsx, 0) {$\cdots$};
    \draw[->] (leftdots) edge (s t-1);
    \draw[->] (s t-1) edge (s t);
    \draw[->] (s t) edge (s t+1);
    \draw[->] (s t+1) edge (rightdots);
    \node[anchor=east] at (-\labelx, -\ysep) {input \big\{};
    \node[anchor=east] at (-\labelx, 0) {stack \big\{};
    \node[anchor=east] at (-\labelx, \ysep) {output \big\{};
\end{tikzpicture}}
    \caption{
        Conceptual diagram of a stack attention sublayer, unrolled across a portion of time. Dotted arrows indicate linear transformations, and dashed arrows indicate residual connections.
    }
    \label{fig:stack-attention-diagram}
\end{figure}

\section{Experimental Details}
\label{sec:experimental-details}

Here, we provide more details about the experiments in \cref{sec:experiments}.

For all models, wherever dropout is applicable, we use a dropout rate of 0.1. In our transformers, all of the scaled dot-product attention sublayers are causally masked and use 4 heads.

Following \citet{dusell-chiang-2024-stack}, we initialize all fully-connected layers with Xavier uniform initialization \citep{glorot-bengio-2010-understanding}, except for layers involved in the recurrent update of RNNs and LSTMs and in the standard scaled dot-product attention layers in transformers. For layer norm, we initialize all weights to 1 and all biases to 0. We initialize all other parameters by sampling uniformly from $[-0.1, 0.1]$.

We group examples of similar lengths into minibatches. We limit the total number of tokens per minibatch, including $\bos$, $\eos$, and padding tokens, by a hyperparameter $\theBatchSize$. For each string $\theString$ in each minibatch, we compute $-\log \theModelProbDist(\theString)$ and take the mean over all $\theString$ in the batch to form the loss function that we minimize. We optimize all models with Adam.\footnote{We found that using AdamW (Adam with weight decay) resulted in generally \emph{weaker}, not stronger, hierarchical generalization.} We use $L^2$ gradient clipping with a threshold of 10. We use cross-entropy of whole strings $\theString$ in the validation set, micro-averaged by the length of $\theString$ plus $\eos$, to control the learning rate schedule and early stopping. We measure this at regular checkpoints, which we take every 80k training examples. We multiply the learning rate by 0.5 whenever the validation cross-entropy does not improve after 2 checkpoints. We stop early when the validation cross-entropy does not improve after 3 checkpoints. When evaluating a model, we always use the parameters corresponding to the checkpoint that resulted in the lowest validation cross-entropy during training.

During our hyperparameter search, we randomly sample $\theBatchSize$ from a uniform distribution over $[512, 2048]$, and we randomly sample the initial learning rate from a log-uniform distribution over $[10^{-5}, 10^{-3}]$.

\section{Full Results}
\label{sec:full-results}

Here, we provide unabridged results for all tasks in \cref{tab:question-formation,tab:question-formation-in-german,tab:tense-reinflection-with-do,tab:passivization,tab:question-formation-with-have,tab:tense-reinflection,tab:passivization-in-german}, including models that did not master the in-distribution test set.

As an additional metric, for both full accuracy and partial accuracy, we report the log-ratio of the hierarchical rule accuracy to the linear rule accuracy. Positive values indicate a preference for the hierarchical rule; and negative, for the linear. Let $\theGeneralizationSet$ be the generalization set according to the hierarchical rule, and let $\theLinearGeneralizationSet$ be the generalization set according to the linear rule. We compute the log-ratio of the full accuracy as $\log(\fullAccuracy{\theModel}{\theGeneralizationSet} / \fullAccuracy{\theModel}{\theLinearGeneralizationSet})$; similarly for partial accuracy.\footnote{This computes the ratio at the model level. An alternative would be to compute the ratio at the \emph{sentence} level, but this would not work for tasks where we need to estimate the expected partial accuracy, since sentence-level estimates can be 0 and cause division by 0. For consistency, we use model-level ratios across the board.}

In all tables, each row is the mean of 5 runs with standard deviations in small text and the best mean value of each column in \textbf{bold}. The meanings of the columns are: ``Test'' = in-distribution test set; ``Hier.'' = hierarchical generalization set; ``Linear'' = linear generalization set; ``Log Ratio'' = log-ratio of the hierarchical generalization accuracy to the linear generalization accuracy, as defined above. The meanings of the model labels are: ``\transformerLabel{}'' = transformer; ``\rnnLabel{}'' = simple RNN; ``\lstmLabel{}'' = LSTM; ``+\supLabel{}'' = with superposition stack; ``+\ndLabel{}'' = with nondeterministic stack; ``+\readingLabel{}'' = with the stack reading short-circuited to the output. We provide comparable results from prior work. For the sake of completeness, we also include results on three additional datasets: the modified English question formation task of \citet{mueller-etal-2022-coloring}, the original tense reinflection task of \citet{mccoy-etal-2020-syntax} without auxiliary \textit{do}, and the German passivization task of \citet{mueller-etal-2022-coloring}.

\begin{table*}
    \centering
    \small
    \begin{tabular}{@{}lS[table-format=1.3]@{}lS[table-format=1.3]@{}lS[table-format=1.3]@{}lS[table-format=-1.3]@{}lS[table-format=1.3]@{}lS[table-format=1.3]@{}lS[table-format=-1.3]@{}l@{}}
\toprule
& \multicolumn{8}{c}{Full Accuracy} & \multicolumn{6}{c}{Partial Accuracy} \\
\cmidrule(lr){2-9} \cmidrule(lr){10-15}
Architecture & \multicolumn{2}{c}{Test $\uparrow$} & \multicolumn{2}{c}{Hier. $\uparrow$} & \multicolumn{2}{c}{Linear $\downarrow$} & \multicolumn{2}{c}{Log Ratio $\uparrow$} & \multicolumn{2}{c}{Hier. $\uparrow$} & \multicolumn{2}{c}{Linear $\downarrow$} & \multicolumn{2}{c}{Log Ratio $\uparrow$} \\
\midrule
\transformerLabel{} & \meanAndVarBold{0.999}{.00} & \meanAndVar{0.005}{.01} & \meanAndVar{0.112}{.06} & \meanAndVar{-4.926}{2.44} & \meanAndVar{0.645}{.25} & \meanAndVar{0.355}{.25} & \meanAndVar{0.757}{1.35} \\
\transformerLabel{}+\supLabel{} & \meanAndVar{0.999}{.00} & \meanAndVar{0.039}{.04} & \meanAndVar{0.202}{.15} & \meanAndVar{-0.365}{3.17} & \meanAndVar{0.325}{.21} & \meanAndVar{0.675}{.21} & \meanAndVar{-1.004}{1.29} \\
\transformerLabel{}+\supLabel{}+\supLabel{} & \meanAndVar{0.998}{.00} & \meanAndVar{0.161}{.18} & \meanAndVar{0.222}{.10} & \meanAndVar{-0.633}{.87} & \meanAndVar{0.697}{.08} & \meanAndVar{0.303}{.08} & \meanAndVar{0.865}{.40} \\
\transformerLabel{}+\ndLabel{} & \meanAndVar{0.999}{.00} & \meanAndVarBold{0.318}{.19} & \meanAndVar{0.018}{.02} & \meanAndVarBold{8.563}{8.11} & \meanAndVar{0.732}{.15} & \meanAndVar{0.268}{.15} & \meanAndVar{2.461}{3.52} \\
\transformerLabel{}+\ndLabel{}+\ndLabel{} & \meanAndVar{0.995}{.00} & \meanAndVar{0.191}{.22} & \meanAndVar{0.067}{.13} & \meanAndVar{7.898}{9.95} & \meanAndVarBold{0.862}{.17} & \meanAndVarBold{0.135}{.16} & \meanAndVarBold{4.049}{2.86} \\
\rnnLabel{} & \meanAndVar{0.465}{.31} & \meanAndVar{0.000}{.00} & \meanAndVar{0.000}{.00} & \meanAndVar{1.372}{3.73} & \meanAndVar{0.659}{.20} & \meanAndVar{0.315}{.20} & \meanAndVar{1.054}{1.26} \\
\rnnLabel{}+\supLabel{} & \meanAndVar{0.022}{.02} & \meanAndVar{0.000}{.00} & \meanAndVar{0.000}{.00} & \meanAndVar{1.255}{3.13} & \meanAndVar{0.417}{.16} & \meanAndVar{0.457}{.06} & \meanAndVar{-0.234}{.50} \\
\rnnLabel{}+\supLabel{}+\readingLabel{} & \meanAndVar{0.011}{.01} & \meanAndVar{0.000}{.00} & \meanAndVar{0.000}{.00} & \meanAndVar{-1.931}{1.70} & \meanAndVar{0.482}{.02} & \meanAndVar{0.494}{.02} & \meanAndVar{-0.024}{.05} \\
\rnnLabel{}+\ndLabel{} & \meanAndVar{0.027}{.02} & \meanAndVar{0.000}{.00} & \meanAndVar{0.000}{.00} & \meanAndVar{-0.806}{2.67} & \meanAndVar{0.525}{.07} & \meanAndVar{0.433}{.07} & \meanAndVar{0.200}{.29} \\
\rnnLabel{}+\ndLabel{}+\readingLabel{} & \meanAndVar{0.019}{.01} & \meanAndVar{0.000}{.00} & \meanAndVarBold{0.000}{.00} & \meanAndVar{0.032}{3.00} & \meanAndVar{0.511}{.03} & \meanAndVar{0.479}{.03} & \meanAndVar{0.064}{.12} \\
\lstmLabel{} & \meanAndVar{0.003}{.00} & \meanAndVar{0.000}{.00} & \meanAndVar{0.000}{.00} & \meanAndVar{5.404}{5.01} & \meanAndVar{0.497}{.00} & \meanAndVar{0.497}{.00} & \meanAndVar{0.000}{.00} \\
\lstmLabel{}+\supLabel{} & \meanAndVar{0.002}{.00} & \meanAndVar{0.000}{.00} & \meanAndVar{0.000}{.00} & \meanAndVar{5.284}{.83} & \meanAndVar{0.498}{.00} & \meanAndVar{0.498}{.00} & \meanAndVar{0.000}{.00} \\
\lstmLabel{}+\supLabel{}+\readingLabel{} & \meanAndVar{0.005}{.00} & \meanAndVar{0.001}{.00} & \meanAndVar{0.000}{.00} & \meanAndVar{3.929}{3.10} & \meanAndVar{0.505}{.02} & \meanAndVar{0.482}{.03} & \meanAndVar{0.049}{.10} \\
\lstmLabel{}+\ndLabel{} & \meanAndVar{0.002}{.00} & \meanAndVar{0.000}{.00} & \meanAndVar{0.000}{.00} & \meanAndVar{4.045}{2.03} & \meanAndVar{0.499}{.00} & \meanAndVar{0.499}{.00} & \meanAndVar{0.001}{.00} \\
\lstmLabel{}+\ndLabel{}+\readingLabel{} & \meanAndVar{0.118}{.17} & \meanAndVar{0.009}{.02} & \meanAndVar{0.000}{.00} & \meanAndVar{5.598}{4.13} & \meanAndVar{0.505}{.07} & \meanAndVar{0.454}{.05} & \meanAndVar{0.104}{.21} \\
\midrule
GRU+loc. attn.\mccoyFootnote{} & 0.77 && \text{--} && \text{--} && \text{--} && 0.78 && \text{--} && \text{--} & \\
ON-LSTM\mccoyFootnote{} & 0.93 && \text{--} && \text{--} && \text{--} && 0.05 && \text{--} && \text{--} & \\
Tree-GRU\mccoyFootnote{}\treeFootnote{} & 0.96 && \text{--} && \text{--} && \text{--} && 0.99 && \text{--} && \text{--} & \\
Tf\murtyFootnote{} & \text{--} && \text{--} && \text{--} && \text{--} && 0.34 && \text{--} && \text{--} & \\
Tf\murtyFootnote{}\overtrainedFootnote{} & \text{--} && \text{--} && \text{--} && \text{--} && 0.92 && \text{--} && \text{--} & \\
Tf\ahujaFootnote{}\overtrainedFootnote{} & 1.00 && \text{--} && \text{--} && \text{--} && 0.75 && \text{--} && \text{--} & \\
Tf\qinFootnote{}\overtrainedFootnote{} & 1.00 && \text{--} && \text{--} && \text{--} && 0.84 && \text{--} && \text{--} & \\
\bottomrule
\end{tabular}

    \caption{\label{tab:question-formation}
        Results on question formation. \mccoyFootnoteText{} \murtyFootnoteText{} \ahujaFootnoteText{} \qinFootnoteText{} \treeFootnoteText{} \overtrainedFootnoteText{}
    }
\end{table*}

\begin{table*}
    \centering
    \small
    \begin{tabular}{@{}lS[table-format=1.3]@{}lS[table-format=1.3]@{}lS[table-format=1.3]@{}lS[table-format=2.3]@{}lS[table-format=1.3]@{}lS[table-format=1.3]@{}lS[table-format=2.3]@{}l@{}}
\toprule
& \multicolumn{8}{c}{Full Accuracy} & \multicolumn{6}{c}{Partial Accuracy} \\
\cmidrule(lr){2-9} \cmidrule(lr){10-15}
Architecture & \multicolumn{2}{c}{Test $\uparrow$} & \multicolumn{2}{c}{Hier. $\uparrow$} & \multicolumn{2}{c}{Linear $\downarrow$} & \multicolumn{2}{c}{Log Ratio $\uparrow$} & \multicolumn{2}{c}{Hier. $\uparrow$} & \multicolumn{2}{c}{Linear $\downarrow$} & \multicolumn{2}{c}{Log Ratio $\uparrow$} \\
\midrule
\transformerLabel{} & \meanAndVar{0.996}{.00} & \meanAndVar{0.000}{.00} & \meanAndVar{0.000}{.00} & \meanAndVar{6.535}{2.62} & \meanAndVar{0.868}{.26} & \meanAndVar{0.003}{.01} & \meanAndVar{7.330}{2.32} \\
\transformerLabel{}+\supLabel{} & \meanAndVar{0.983}{.00} & \meanAndVar{0.000}{.00} & \meanAndVar{0.000}{.00} & \meanAndVar{7.458}{3.57} & \meanAndVar{0.999}{.00} & \meanAndVar{0.001}{.00} & \meanAndVar{7.251}{.75} \\
\transformerLabel{}+\supLabel{}+\supLabel{} & \meanAndVar{0.999}{.00} & \meanAndVar{0.000}{.00} & \meanAndVar{0.000}{.00} & \meanAndVar{12.645}{6.15} & \meanAndVar{0.996}{.01} & \meanAndVar{0.004}{.01} & \meanAndVar{8.688}{2.53} \\
\transformerLabel{}+\ndLabel{} & \meanAndVarBold{0.999}{.00} & \meanAndVar{0.004}{.01} & \meanAndVar{0.000}{.00} & \meanAndVar{20.631}{8.71} & \meanAndVarBold{1.000}{.00} & \meanAndVarBold{0.000}{.00} & \meanAndVarBold{9.810}{.47} \\
\transformerLabel{}+\ndLabel{}+\ndLabel{} & \meanAndVar{0.997}{.00} & \meanAndVar{0.001}{.00} & \meanAndVar{0.000}{.00} & \meanAndVar{16.850}{10.09} & \meanAndVar{0.979}{.04} & \meanAndVar{0.000}{.00} & \meanAndVar{8.484}{.97} \\
\rnnLabel{} & \meanAndVar{0.324}{.12} & \meanAndVar{0.000}{.00} & \meanAndVar{0.000}{.00} & \meanAndVar{9.445}{3.23} & \meanAndVar{0.845}{.08} & \meanAndVar{0.110}{.03} & \meanAndVar{2.063}{.33} \\
\rnnLabel{}+\supLabel{} & \meanAndVar{0.274}{.12} & \meanAndVar{0.000}{.00} & \meanAndVar{0.000}{.00} & \meanAndVar{10.984}{4.60} & \meanAndVar{0.952}{.02} & \meanAndVar{0.025}{.01} & \meanAndVar{3.715}{.43} \\
\rnnLabel{}+\supLabel{}+\readingLabel{} & \meanAndVar{0.210}{.04} & \meanAndVar{0.000}{.00} & \meanAndVarBold{0.000}{.00} & \meanAndVar{13.720}{4.41} & \meanAndVar{0.965}{.03} & \meanAndVar{0.011}{.01} & \meanAndVar{5.096}{1.06} \\
\rnnLabel{}+\ndLabel{} & \meanAndVar{0.189}{.02} & \meanAndVar{0.000}{.00} & \meanAndVar{0.000}{.00} & \meanAndVar{7.267}{3.87} & \meanAndVar{0.854}{.11} & \meanAndVar{0.125}{.11} & \meanAndVar{2.582}{1.47} \\
\rnnLabel{}+\ndLabel{}+\readingLabel{} & \meanAndVar{0.153}{.04} & \meanAndVar{0.000}{.00} & \meanAndVar{0.000}{.00} & \meanAndVar{10.619}{5.22} & \meanAndVar{0.892}{.06} & \meanAndVar{0.091}{.06} & \meanAndVar{2.575}{.91} \\
\lstmLabel{} & \meanAndVar{0.234}{.10} & \meanAndVarBold{0.012}{.01} & \meanAndVar{0.000}{.00} & \meanAndVar{22.493}{3.06} & \meanAndVar{0.992}{.01} & \meanAndVar{0.003}{.00} & \meanAndVar{6.823}{1.66} \\
\lstmLabel{}+\supLabel{} & \meanAndVar{0.200}{.15} & \meanAndVar{0.007}{.01} & \meanAndVar{0.000}{.00} & \meanAndVar{18.428}{6.10} & \meanAndVar{0.980}{.02} & \meanAndVar{0.010}{.01} & \meanAndVar{5.605}{1.82} \\
\lstmLabel{}+\supLabel{}+\readingLabel{} & \meanAndVar{0.356}{.10} & \meanAndVar{0.002}{.00} & \meanAndVar{0.000}{.00} & \meanAndVar{22.105}{7.12} & \meanAndVar{0.918}{.15} & \meanAndVar{0.078}{.15} & \meanAndVar{6.396}{3.62} \\
\lstmLabel{}+\ndLabel{} & \meanAndVar{0.199}{.04} & \meanAndVar{0.000}{.00} & \meanAndVar{0.000}{.00} & \meanAndVar{22.320}{5.27} & \meanAndVar{0.960}{.06} & \meanAndVar{0.003}{.00} & \meanAndVar{6.758}{1.71} \\
\lstmLabel{}+\ndLabel{}+\readingLabel{} & \meanAndVar{0.648}{.24} & \meanAndVar{0.008}{.01} & \meanAndVar{0.000}{.00} & \meanAndVarBold{23.347}{4.98} & \meanAndVar{0.887}{.22} & \meanAndVar{0.000}{.00} & \meanAndVar{8.862}{.95} \\
\midrule
mT5\muellerFootnote{}\pretrainedFootnote{} & 1.00 && \text{--} && \text{--} && \text{--} && 1.00 && \text{--} && \text{--} & \\
mBART\muellerFootnote{}\pretrainedFootnote{} & 1.00 && \text{--} && \text{--} && \text{--} && 0.82 && \text{--} && \text{--} & \\
Tf\ahujaFootnote{}\overtrainedFootnote{} & 1.00 && \text{--} && \text{--} && \text{--} && 0.95 && \text{--} && \text{--} & \\
\bottomrule
\end{tabular}

    \caption{\label{tab:question-formation-in-german}
        Results on German question formation. \muellerFootnoteText{} \ahujaFootnoteText{} \pretrainedFootnoteText{} \overtrainedFootnoteText{}
    }
\end{table*}

\begin{table*}
    \centering
    \small
    \begin{tabular}{@{}lS[table-format=1.3]@{}lS[table-format=1.3]@{}lS[table-format=1.3]@{}lS[table-format=-1.3]@{}lS[table-format=1.3]@{}lS[table-format=1.3]@{}lS[table-format=-1.3]@{}l@{}}
\toprule
& \multicolumn{8}{c}{Full Accuracy} & \multicolumn{6}{c}{Partial Accuracy} \\
\cmidrule(lr){2-9} \cmidrule(lr){10-15}
Architecture & \multicolumn{2}{c}{Test $\uparrow$} & \multicolumn{2}{c}{Hier. $\uparrow$} & \multicolumn{2}{c}{Linear $\downarrow$} & \multicolumn{2}{c}{Log Ratio $\uparrow$} & \multicolumn{2}{c}{Hier. $\uparrow$} & \multicolumn{2}{c}{Linear $\downarrow$} & \multicolumn{2}{c}{Log Ratio $\uparrow$} \\
\midrule
\transformerLabel{} & \meanAndVar{0.998}{.00} & \meanAndVar{0.145}{.15} & \meanAndVar{0.852}{.14} & \meanAndVar{-2.408}{1.39} & \meanAndVar{0.145}{.15} & \meanAndVar{0.854}{.15} & \meanAndVar{-2.406}{1.39} \\
\transformerLabel{}+\supLabel{} & \meanAndVar{0.998}{.00} & \meanAndVar{0.368}{.10} & \meanAndVar{0.629}{.10} & \meanAndVar{-0.560}{.43} & \meanAndVar{0.370}{.10} & \meanAndVar{0.630}{.10} & \meanAndVar{-0.557}{.43} \\
\transformerLabel{}+\supLabel{}+\supLabel{} & \meanAndVarBold{0.999}{.00} & \meanAndVar{0.127}{.12} & \meanAndVar{0.871}{.12} & \meanAndVar{-2.531}{1.34} & \meanAndVar{0.128}{.12} & \meanAndVar{0.871}{.12} & \meanAndVar{-2.529}{1.34} \\
\transformerLabel{}+\ndLabel{} & \meanAndVar{0.998}{.00} & \meanAndVarBold{0.416}{.10} & \meanAndVar{0.521}{.14} & \meanAndVarBold{-0.210}{.55} & \meanAndVarBold{0.476}{.15} & \meanAndVar{0.523}{.15} & \meanAndVarBold{-0.091}{.62} \\
\transformerLabel{}+\ndLabel{}+\ndLabel{} & \meanAndVar{0.995}{.01} & \meanAndVar{0.165}{.11} & \meanAndVar{0.829}{.12} & \meanAndVar{-3.246}{3.70} & \meanAndVar{0.167}{.12} & \meanAndVar{0.831}{.12} & \meanAndVar{-3.427}{4.07} \\
\rnnLabel{} & \meanAndVar{0.508}{.19} & \meanAndVar{0.005}{.00} & \meanAndVar{0.432}{.18} & \meanAndVar{-4.895}{1.22} & \meanAndVar{0.039}{.01} & \meanAndVar{0.918}{.03} & \meanAndVar{-3.251}{.49} \\
\rnnLabel{}+\supLabel{} & \meanAndVar{0.129}{.05} & \meanAndVar{0.000}{.00} & \meanAndVar{0.080}{.04} & \meanAndVar{-5.653}{.62} & \meanAndVar{0.041}{.00} & \meanAndVar{0.912}{.01} & \meanAndVar{-3.106}{.11} \\
\rnnLabel{}+\supLabel{}+\readingLabel{} & \meanAndVar{0.233}{.15} & \meanAndVar{0.001}{.00} & \meanAndVar{0.177}{.13} & \meanAndVar{-4.893}{.32} & \meanAndVar{0.044}{.01} & \meanAndVar{0.891}{.02} & \meanAndVar{-3.024}{.17} \\
\rnnLabel{}+\ndLabel{} & \meanAndVar{0.182}{.11} & \meanAndVar{0.000}{.00} & \meanAndVar{0.143}{.10} & \meanAndVar{-6.013}{.94} & \meanAndVar{0.031}{.01} & \meanAndVar{0.934}{.02} & \meanAndVar{-3.448}{.34} \\
\rnnLabel{}+\ndLabel{}+\readingLabel{} & \meanAndVar{0.105}{.09} & \meanAndVar{0.001}{.00} & \meanAndVar{0.070}{.08} & \meanAndVar{-4.645}{1.40} & \meanAndVar{0.063}{.02} & \meanAndVar{0.905}{.02} & \meanAndVar{-2.701}{.33} \\
\lstmLabel{} & \meanAndVar{0.010}{.00} & \meanAndVar{0.000}{.00} & \meanAndVar{0.002}{.00} & \meanAndVar{-4.520}{2.54} & \meanAndVar{0.036}{.05} & \meanAndVar{0.463}{.05} & \meanAndVar{-4.094}{2.34} \\
\lstmLabel{}+\supLabel{} & \meanAndVar{0.011}{.00} & \meanAndVar{0.000}{.00} & \meanAndVar{0.003}{.00} & \meanAndVar{-3.575}{1.50} & \meanAndVar{0.038}{.04} & \meanAndVarBold{0.431}{.07} & \meanAndVar{-2.964}{1.24} \\
\lstmLabel{}+\supLabel{}+\readingLabel{} & \meanAndVar{0.008}{.00} & \meanAndVar{0.000}{.00} & \meanAndVar{0.002}{.00} & \meanAndVar{-3.899}{1.36} & \meanAndVar{0.039}{.03} & \meanAndVar{0.459}{.03} & \meanAndVar{-2.921}{1.07} \\
\lstmLabel{}+\ndLabel{} & \meanAndVar{0.009}{.00} & \meanAndVar{0.000}{.00} & \meanAndVarBold{0.001}{.00} & \meanAndVar{-6.612}{1.11} & \meanAndVar{0.007}{.01} & \meanAndVar{0.492}{.01} & \meanAndVar{-5.025}{1.50} \\
\lstmLabel{}+\ndLabel{}+\readingLabel{} & \meanAndVar{0.045}{.03} & \meanAndVar{0.000}{.00} & \meanAndVar{0.022}{.02} & \meanAndVar{-4.986}{1.35} & \meanAndVar{0.029}{.02} & \meanAndVar{0.868}{.20} & \meanAndVar{-3.675}{.92} \\
\bottomrule
\end{tabular}

    \caption{\label{tab:tense-reinflection-with-do}
        Results on the tense reinflection task with auxiliary \textit{do}.
    }
\end{table*}

\begin{table*}
    \centering
    \small
    \begin{tabular}{@{}lS[table-format=1.3]@{}lS[table-format=1.3]@{}lS[table-format=1.3]@{}lS[table-format=-2.3]@{}lS[table-format=1.3]@{}lS[table-format=1.3]@{}lS[table-format=-1.3]@{}l@{}}
\toprule
& \multicolumn{8}{c}{Full Accuracy} & \multicolumn{6}{c}{Partial Accuracy} \\
\cmidrule(lr){2-9} \cmidrule(lr){10-15}
Architecture & \multicolumn{2}{c}{Test $\uparrow$} & \multicolumn{2}{c}{Hier. $\uparrow$} & \multicolumn{2}{c}{Linear $\downarrow$} & \multicolumn{2}{c}{Log Ratio $\uparrow$} & \multicolumn{2}{c}{Hier. $\uparrow$} & \multicolumn{2}{c}{Linear $\downarrow$} & \multicolumn{2}{c}{Log Ratio $\uparrow$} \\
\midrule
\transformerLabel{} & \meanAndVar{0.999}{.00} & \meanAndVar{0.000}{.00} & \meanAndVar{0.000}{.00} & \meanAndVar{-19.383}{9.52} & \meanAndVar{0.526}{.34} & \meanAndVar{0.196}{.18} & \meanAndVar{1.075}{2.13} \\
\transformerLabel{}+\supLabel{} & \meanAndVar{0.998}{.00} & \meanAndVar{0.000}{.00} & \meanAndVar{0.000}{.00} & \meanAndVar{-19.484}{5.88} & \meanAndVar{0.565}{.31} & \meanAndVar{0.144}{.19} & \meanAndVarBold{1.555}{2.09} \\
\transformerLabel{}+\supLabel{}+\supLabel{} & \meanAndVar{0.998}{.00} & \meanAndVar{0.000}{.00} & \meanAndVar{0.001}{.00} & \meanAndVar{-13.042}{12.01} & \meanAndVarBold{0.758}{.37} & \meanAndVar{0.276}{.37} & \meanAndVar{1.511}{2.54} \\
\transformerLabel{}+\ndLabel{} & \meanAndVar{0.988}{.01} & \meanAndVar{0.000}{.00} & \meanAndVar{0.000}{.00} & \meanAndVar{-19.543}{9.56} & \meanAndVar{0.612}{.46} & \meanAndVar{0.383}{.42} & \meanAndVar{0.717}{3.01} \\
\transformerLabel{}+\ndLabel{}+\ndLabel{} & \meanAndVarBold{1.000}{.00} & \meanAndVarBold{0.000}{.00} & \meanAndVar{0.000}{.00} & \meanAndVar{-8.729}{5.86} & \meanAndVar{0.569}{.38} & \meanAndVar{0.280}{.36} & \meanAndVar{1.188}{2.45} \\
\rnnLabel{} & \meanAndVar{0.926}{.02} & \meanAndVar{0.000}{.00} & \meanAndVar{0.000}{.00} & \meanAndVar{-16.376}{3.19} & \meanAndVar{0.061}{.01} & \meanAndVar{0.146}{.02} & \meanAndVar{-0.881}{.21} \\
\rnnLabel{}+\supLabel{} & \meanAndVar{0.848}{.11} & \meanAndVar{0.000}{.00} & \meanAndVar{0.000}{.00} & \meanAndVar{-17.820}{4.36} & \meanAndVar{0.056}{.01} & \meanAndVar{0.102}{.02} & \meanAndVar{-0.567}{.19} \\
\rnnLabel{}+\supLabel{}+\readingLabel{} & \meanAndVar{0.537}{.42} & \meanAndVar{0.000}{.00} & \meanAndVar{0.000}{.00} & \meanAndVar{-19.488}{8.70} & \meanAndVar{0.048}{.00} & \meanAndVar{0.079}{.02} & \meanAndVar{-0.459}{.24} \\
\rnnLabel{}+\ndLabel{} & \meanAndVar{0.747}{.24} & \meanAndVar{0.000}{.00} & \meanAndVar{0.000}{.00} & \meanAndVar{-15.348}{5.85} & \meanAndVar{0.047}{.00} & \meanAndVar{0.082}{.01} & \meanAndVar{-0.550}{.13} \\
\rnnLabel{}+\ndLabel{}+\readingLabel{} & \meanAndVar{0.669}{.34} & \meanAndVar{0.000}{.00} & \meanAndVar{0.000}{.00} & \meanAndVar{-12.230}{6.20} & \meanAndVar{0.057}{.01} & \meanAndVar{0.108}{.05} & \meanAndVar{-0.564}{.25} \\
\lstmLabel{} & \meanAndVar{0.007}{.01} & \meanAndVar{0.000}{.00} & \meanAndVar{0.000}{.00} & \meanAndVar{4.900}{13.58} & \meanAndVar{0.057}{.02} & \meanAndVar{0.043}{.01} & \meanAndVar{0.240}{.29} \\
\lstmLabel{}+\supLabel{} & \meanAndVar{0.009}{.00} & \meanAndVar{0.000}{.00} & \meanAndVar{0.000}{.00} & \meanAndVar{-8.372}{8.56} & \meanAndVar{0.089}{.03} & \meanAndVar{0.059}{.02} & \meanAndVar{0.411}{.45} \\
\lstmLabel{}+\supLabel{}+\readingLabel{} & \meanAndVar{0.011}{.00} & \meanAndVar{0.000}{.00} & \meanAndVarBold{0.000}{.00} & \meanAndVarBold{17.250}{21.00} & \meanAndVar{0.071}{.02} & \meanAndVar{0.062}{.03} & \meanAndVar{0.211}{.26} \\
\lstmLabel{}+\ndLabel{} & \meanAndVar{0.004}{.00} & \meanAndVar{0.000}{.00} & \meanAndVar{0.000}{.00} & \meanAndVar{14.020}{14.82} & \meanAndVar{0.056}{.02} & \meanAndVarBold{0.042}{.01} & \meanAndVar{0.252}{.29} \\
\lstmLabel{}+\ndLabel{}+\readingLabel{} & \meanAndVar{0.209}{.39} & \meanAndVar{0.000}{.00} & \meanAndVar{0.000}{.00} & \meanAndVar{-1.055}{10.84} & \meanAndVar{0.184}{.10} & \meanAndVar{0.048}{.01} & \meanAndVar{1.218}{.54} \\
\midrule
T5\muellerFootnote{}\pretrainedFootnote{} & 1.00 && \text{--} && \text{--} && \text{--} && 1.00 && \text{--} && \text{--} & \\
mT5\muellerFootnote{}\pretrainedFootnote{} & 1.00 && \text{--} && \text{--} && \text{--} && 1.00 && \text{--} && \text{--} & \\
BART\muellerFootnote{}\pretrainedFootnote{} & 0.95 && \text{--} && \text{--} && \text{--} && 1.00 && \text{--} && \text{--} & \\
mBART\muellerFootnote{}\pretrainedFootnote{} & 1.00 && \text{--} && \text{--} && \text{--} && 0.80 && \text{--} && \text{--} & \\
Tf\ahujaFootnote{}\overtrainedFootnote{} & 1.00 && \text{--} && \text{--} && \text{--} && 1.00 && \text{--} && \text{--} & \\
\bottomrule
\end{tabular}

    \caption{\label{tab:passivization}
        Results on the passivization task. \muellerFootnoteText{} \ahujaFootnoteText{} \pretrainedFootnoteText{} \overtrainedFootnoteText{} Note that the hierarchical and linear partial accuracy scores can sum to more than 1, because the generalization set includes examples where both rules move the same noun type.
    }
\end{table*}

\begin{table*}
    \centering
    \small
    \begin{tabular}{@{}lS[table-format=1.3]@{}lS[table-format=1.3]@{}lS[table-format=1.3]@{}lS[table-format=-1.3]@{}lS[table-format=1.3]@{}lS[table-format=1.3]@{}lS[table-format=-1.3]@{}l@{}}
\toprule
& \multicolumn{8}{c}{Full Accuracy} & \multicolumn{6}{c}{Partial Accuracy} \\
\cmidrule(lr){2-9} \cmidrule(lr){10-15}
Architecture & \multicolumn{2}{c}{Test $\uparrow$} & \multicolumn{2}{c}{Hier. $\uparrow$} & \multicolumn{2}{c}{Linear $\downarrow$} & \multicolumn{2}{c}{Log Ratio $\uparrow$} & \multicolumn{2}{c}{Hier. $\uparrow$} & \multicolumn{2}{c}{Linear $\downarrow$} & \multicolumn{2}{c}{Log Ratio $\uparrow$} \\
\midrule
\transformerLabel{} & \meanAndVar{0.993}{.01} & \meanAndVar{0.006}{.01} & \meanAndVar{0.325}{.01} & \meanAndVar{-5.253}{1.60} & \meanAndVar{0.382}{.18} & \meanAndVar{0.618}{.18} & \meanAndVar{-0.604}{.91} \\
\transformerLabel{}+\supLabel{} & \meanAndVar{0.998}{.00} & \meanAndVar{0.025}{.02} & \meanAndVar{0.213}{.12} & \meanAndVar{-3.962}{3.70} & \meanAndVar{0.669}{.20} & \meanAndVar{0.331}{.20} & \meanAndVar{1.361}{2.03} \\
\transformerLabel{}+\supLabel{}+\supLabel{} & \meanAndVarBold{1.000}{.00} & \meanAndVar{0.074}{.08} & \meanAndVar{0.187}{.11} & \meanAndVar{-2.945}{3.23} & \meanAndVar{0.699}{.15} & \meanAndVar{0.301}{.15} & \meanAndVar{1.092}{1.10} \\
\transformerLabel{}+\ndLabel{} & \meanAndVar{0.995}{.00} & \meanAndVar{0.140}{.16} & \meanAndVar{0.129}{.12} & \meanAndVar{3.479}{9.70} & \meanAndVar{0.523}{.32} & \meanAndVar{0.477}{.32} & \meanAndVar{1.261}{3.64} \\
\transformerLabel{}+\ndLabel{}+\ndLabel{} & \meanAndVar{0.989}{.01} & \meanAndVarBold{0.190}{.20} & \meanAndVar{0.006}{.01} & \meanAndVarBold{7.720}{6.34} & \meanAndVarBold{0.756}{.13} & \meanAndVarBold{0.231}{.12} & \meanAndVarBold{2.120}{2.47} \\
\rnnLabel{} & \meanAndVar{0.147}{.08} & \meanAndVar{0.000}{.00} & \meanAndVar{0.000}{.00} & \meanAndVar{-0.421}{1.54} & \meanAndVar{0.487}{.01} & \meanAndVar{0.487}{.01} & \meanAndVar{-0.000}{.00} \\
\rnnLabel{}+\supLabel{} & \meanAndVar{0.044}{.03} & \meanAndVar{0.000}{.00} & \meanAndVar{0.000}{.00} & \meanAndVar{-1.058}{2.92} & \meanAndVar{0.502}{.05} & \meanAndVar{0.447}{.06} & \meanAndVar{0.122}{.24} \\
\rnnLabel{}+\supLabel{}+\readingLabel{} & \meanAndVar{0.023}{.03} & \meanAndVar{0.000}{.00} & \meanAndVarBold{0.000}{.00} & \meanAndVar{0.703}{1.21} & \meanAndVar{0.441}{.09} & \meanAndVar{0.442}{.09} & \meanAndVar{-0.000}{.00} \\
\rnnLabel{}+\ndLabel{} & \meanAndVar{0.054}{.05} & \meanAndVar{0.000}{.00} & \meanAndVar{0.000}{.00} & \meanAndVar{1.238}{2.17} & \meanAndVar{0.464}{.04} & \meanAndVar{0.464}{.04} & \meanAndVar{-0.000}{.00} \\
\rnnLabel{}+\ndLabel{}+\readingLabel{} & \meanAndVar{0.030}{.02} & \meanAndVar{0.000}{.00} & \meanAndVar{0.000}{.00} & \meanAndVar{-0.221}{4.24} & \meanAndVar{0.494}{.01} & \meanAndVar{0.494}{.01} & \meanAndVar{-0.000}{.00} \\
\lstmLabel{} & \meanAndVar{0.005}{.00} & \meanAndVar{0.000}{.00} & \meanAndVar{0.000}{.00} & \meanAndVar{7.694}{4.26} & \meanAndVar{0.498}{.00} & \meanAndVar{0.498}{.00} & \meanAndVar{-0.000}{.00} \\
\lstmLabel{}+\supLabel{} & \meanAndVar{0.005}{.00} & \meanAndVar{0.000}{.00} & \meanAndVar{0.000}{.00} & \meanAndVar{6.155}{1.56} & \meanAndVar{0.496}{.00} & \meanAndVar{0.496}{.00} & \meanAndVar{-0.001}{.00} \\
\lstmLabel{}+\supLabel{}+\readingLabel{} & \meanAndVar{0.012}{.01} & \meanAndVar{0.002}{.00} & \meanAndVar{0.000}{.00} & \meanAndVar{6.107}{4.63} & \meanAndVar{0.589}{.20} & \meanAndVar{0.390}{.20} & \meanAndVar{1.315}{2.63} \\
\lstmLabel{}+\ndLabel{} & \meanAndVar{0.005}{.00} & \meanAndVar{0.000}{.00} & \meanAndVar{0.000}{.00} & \meanAndVar{4.364}{4.63} & \meanAndVar{0.489}{.01} & \meanAndVar{0.489}{.01} & \meanAndVar{-0.000}{.00} \\
\lstmLabel{}+\ndLabel{}+\readingLabel{} & \meanAndVar{0.013}{.01} & \meanAndVar{0.000}{.00} & \meanAndVar{0.000}{.00} & \meanAndVar{2.680}{4.28} & \meanAndVar{0.419}{.09} & \meanAndVar{0.447}{.11} & \meanAndVar{-0.058}{.34} \\
\midrule
T5\muellerFootnote{}\pretrainedFootnote{} & 1.00 && \text{--} && \text{--} && \text{--} && 0.87 && \text{--} && \text{--} & \\
mT5\muellerFootnote{}\pretrainedFootnote{} & 1.00 && \text{--} && \text{--} && \text{--} && 0.99 && \text{--} && \text{--} & \\
BART\muellerFootnote{}\pretrainedFootnote{} & 0.96 && \text{--} && \text{--} && \text{--} && 0.96 && \text{--} && \text{--} & \\
mBART\muellerFootnote{}\pretrainedFootnote{} & 1.00 && \text{--} && \text{--} && \text{--} && 0.59 && \text{--} && \text{--} & \\
\bottomrule
\end{tabular}

    \caption{\label{tab:question-formation-with-have}
        Results on the modified question formation task of \citet{mueller-etal-2022-coloring} with auxiliary \textit{have}. \muellerFootnoteText{} \pretrainedFootnoteText{}
    }
\end{table*}

\begin{table*}
    \centering
    \small
    \begin{tabular}{@{}lS[table-format=1.3]@{}lS[table-format=1.3]@{}lS[table-format=1.3]@{}lS[table-format=-1.3]@{}lS[table-format=1.3]@{}lS[table-format=1.3]@{}lS[table-format=-1.3]@{}l@{}}
\toprule
& \multicolumn{8}{c}{Full Accuracy} & \multicolumn{6}{c}{Partial Accuracy} \\
\cmidrule(lr){2-9} \cmidrule(lr){10-15}
Architecture & \multicolumn{2}{c}{Test $\uparrow$} & \multicolumn{2}{c}{Hier. $\uparrow$} & \multicolumn{2}{c}{Linear $\downarrow$} & \multicolumn{2}{c}{Log Ratio $\uparrow$} & \multicolumn{2}{c}{Hier. $\uparrow$} & \multicolumn{2}{c}{Linear $\downarrow$} & \multicolumn{2}{c}{Log Ratio $\uparrow$} \\
\midrule
\transformerLabel{} & \meanAndVarBold{0.999}{.00} & \meanAndVar{0.079}{.06} & \meanAndVar{0.905}{.08} & \meanAndVar{-2.814}{1.03} & \meanAndVar{0.086}{.07} & \meanAndVar{0.912}{.07} & \meanAndVar{-2.766}{1.05} \\
\transformerLabel{}+\supLabel{} & \meanAndVar{0.993}{.01} & \meanAndVar{0.036}{.03} & \meanAndVar{0.941}{.04} & \meanAndVar{-3.628}{1.01} & \meanAndVar{0.048}{.03} & \meanAndVar{0.947}{.04} & \meanAndVar{-3.392}{1.08} \\
\transformerLabel{}+\supLabel{}+\supLabel{} & \meanAndVar{0.984}{.01} & \meanAndVar{0.059}{.04} & \meanAndVar{0.921}{.03} & \meanAndVar{-3.126}{1.03} & \meanAndVar{0.061}{.04} & \meanAndVar{0.931}{.03} & \meanAndVar{-3.054}{.97} \\
\transformerLabel{}+\ndLabel{} & \meanAndVar{0.998}{.00} & \meanAndVar{0.056}{.06} & \meanAndVar{0.936}{.07} & \meanAndVar{-4.027}{1.80} & \meanAndVar{0.062}{.07} & \meanAndVar{0.937}{.07} & \meanAndVar{-3.927}{1.80} \\
\transformerLabel{}+\ndLabel{}+\ndLabel{} & \meanAndVar{0.997}{.00} & \meanAndVarBold{0.086}{.09} & \meanAndVar{0.896}{.11} & \meanAndVar{-4.443}{3.16} & \meanAndVar{0.101}{.11} & \meanAndVar{0.897}{.11} & \meanAndVar{-4.457}{3.35} \\
\rnnLabel{} & \meanAndVar{0.051}{.01} & \meanAndVar{0.000}{.00} & \meanAndVar{0.017}{.01} & \meanAndVar{-4.918}{.72} & \meanAndVar{0.012}{.01} & \meanAndVar{0.333}{.03} & \meanAndVar{-3.494}{.64} \\
\rnnLabel{}+\supLabel{} & \meanAndVar{0.024}{.01} & \meanAndVar{0.000}{.00} & \meanAndVar{0.013}{.01} & \meanAndVar{-4.834}{1.15} & \meanAndVar{0.015}{.01} & \meanAndVar{0.414}{.22} & \meanAndVar{-3.716}{.70} \\
\rnnLabel{}+\supLabel{}+\readingLabel{} & \meanAndVar{0.017}{.01} & \meanAndVar{0.000}{.00} & \meanAndVarBold{0.003}{.00} & \meanAndVarBold{-2.118}{2.23} & \meanAndVarBold{0.331}{.40} & \meanAndVarBold{0.247}{.12} & \meanAndVarBold{-0.765}{2.84} \\
\rnnLabel{}+\ndLabel{} & \meanAndVar{0.059}{.03} & \meanAndVar{0.000}{.00} & \meanAndVar{0.032}{.02} & \meanAndVar{-5.441}{.79} & \meanAndVar{0.028}{.01} & \meanAndVar{0.916}{.01} & \meanAndVar{-3.642}{.58} \\
\rnnLabel{}+\ndLabel{}+\readingLabel{} & \meanAndVar{0.024}{.02} & \meanAndVar{0.000}{.00} & \meanAndVar{0.004}{.00} & \meanAndVar{-4.608}{1.67} & \meanAndVar{0.037}{.04} & \meanAndVar{0.544}{.32} & \meanAndVar{-3.061}{1.19} \\
\lstmLabel{} & \meanAndVar{0.029}{.03} & \meanAndVar{0.000}{.00} & \meanAndVar{0.013}{.02} & \meanAndVar{-3.962}{1.17} & \meanAndVar{0.067}{.08} & \meanAndVar{0.931}{.08} & \meanAndVar{-3.128}{.99} \\
\lstmLabel{}+\supLabel{} & \meanAndVar{0.016}{.00} & \meanAndVar{0.000}{.00} & \meanAndVar{0.003}{.00} & \meanAndVar{-4.348}{2.05} & \meanAndVar{0.041}{.06} & \meanAndVar{0.559}{.19} & \meanAndVar{-3.843}{1.93} \\
\lstmLabel{}+\supLabel{}+\readingLabel{} & \meanAndVar{0.018}{.01} & \meanAndVar{0.000}{.00} & \meanAndVar{0.006}{.00} & \meanAndVar{-3.949}{1.25} & \meanAndVar{0.050}{.07} & \meanAndVar{0.943}{.06} & \meanAndVar{-3.696}{1.28} \\
\lstmLabel{}+\ndLabel{} & \meanAndVar{0.012}{.01} & \meanAndVar{0.000}{.00} & \meanAndVar{0.004}{.00} & \meanAndVar{-2.273}{2.13} & \meanAndVar{0.243}{.21} & \meanAndVar{0.746}{.22} & \meanAndVar{-1.926}{1.94} \\
\lstmLabel{}+\ndLabel{}+\readingLabel{} & \meanAndVar{0.017}{.01} & \meanAndVar{0.000}{.00} & \meanAndVar{0.005}{.00} & \meanAndVar{-4.240}{1.36} & \meanAndVar{0.086}{.06} & \meanAndVar{0.889}{.09} & \meanAndVar{-2.978}{1.55} \\
\midrule
LSTM\mccoyFootnote{} & 0.96 && \text{--} && \text{--} && \text{--} && 0.04 && \text{--} && \text{--} & \\
ON-LSTM\mccoyFootnote{} & 0.95 && \text{--} && \text{--} && \text{--} && 0.05 && \text{--} && \text{--} & \\
Tree-GRU\mccoyFootnote{}\treeFootnote{} & 0.96 && \text{--} && \text{--} && \text{--} && 0.94 && \text{--} && \text{--} & \\
Tf\murtyFootnote{} & \text{--} && \text{--} && \text{--} && \text{--} && 0.46 && \text{--} && \text{--} & \\
Tf\murtyFootnote{}\overtrainedFootnote{} & \text{--} && \text{--} && \text{--} && \text{--} && 0.83 && \text{--} && \text{--} & \\
Tf\ahujaFootnote{}\overtrainedFootnote{} & 1.00 && \text{--} && \text{--} && \text{--} && 0.75 && \text{--} && \text{--} & \\
Tf\qinFootnote{}\overtrainedFootnote{} & 1.00 && \text{--} && \text{--} && \text{--} && 0.84 && \text{--} && \text{--} & \\
\bottomrule
\end{tabular}

    \caption{\label{tab:tense-reinflection}
        Results on the original tense reinflection task of \citet{mccoy-etal-2020-syntax} without auxiliary \textit{do}. \mccoyFootnoteText{} \murtyFootnoteText{} \ahujaFootnoteText{} \qinFootnoteText{} \treeFootnoteText{} \overtrainedFootnoteText{}
    }
\end{table*}

\begin{table*}
    \centering
    \small
    \begin{tabular}{@{}lS[table-format=1.3]@{}lS[table-format=1.3]@{}lS[table-format=1.3]@{}lS[table-format=-2.3]@{}lS[table-format=1.3]@{}lS[table-format=1.3]@{}lS[table-format=-1.3]@{}l@{}}
\toprule
& \multicolumn{8}{c}{Full Accuracy} & \multicolumn{6}{c}{Partial Accuracy} \\
\cmidrule(lr){2-9} \cmidrule(lr){10-15}
Architecture & \multicolumn{2}{c}{Test $\uparrow$} & \multicolumn{2}{c}{Hier. $\uparrow$} & \multicolumn{2}{c}{Linear $\downarrow$} & \multicolumn{2}{c}{Log Ratio $\uparrow$} & \multicolumn{2}{c}{Hier. $\uparrow$} & \multicolumn{2}{c}{Linear $\downarrow$} & \multicolumn{2}{c}{Log Ratio $\uparrow$} \\
\midrule
\transformerLabel{} & \meanAndVar{0.991}{.00} & \meanAndVar{0.000}{.00} & \meanAndVar{0.000}{.00} & \meanAndVar{-0.037}{5.49} & \meanAndVar{0.997}{.00} & \meanAndVar{0.058}{.00} & \meanAndVar{2.849}{.00} \\
\transformerLabel{}+\supLabel{} & \meanAndVarBold{0.998}{.00} & \meanAndVar{0.000}{.00} & \meanAndVar{0.000}{.00} & \meanAndVar{0.240}{5.60} & \meanAndVar{0.987}{.02} & \meanAndVar{0.061}{.00} & \meanAndVar{2.791}{.08} \\
\transformerLabel{}+\supLabel{}+\supLabel{} & \meanAndVar{0.997}{.00} & \meanAndVarBold{0.002}{.00} & \meanAndVar{0.000}{.00} & \meanAndVar{1.276}{5.85} & \meanAndVarBold{0.999}{.00} & \meanAndVar{0.058}{.00} & \meanAndVarBold{2.850}{.00} \\
\transformerLabel{}+\ndLabel{} & \meanAndVar{0.994}{.00} & \meanAndVar{0.000}{.00} & \meanAndVar{0.000}{.00} & \meanAndVar{4.321}{3.39} & \meanAndVar{0.956}{.08} & \meanAndVar{0.062}{.01} & \meanAndVar{2.743}{.21} \\
\transformerLabel{}+\ndLabel{}+\ndLabel{} & \meanAndVar{0.997}{.00} & \meanAndVar{0.000}{.00} & \meanAndVar{0.000}{.00} & \meanAndVar{-2.964}{10.00} & \meanAndVar{0.809}{.24} & \meanAndVarBold{0.058}{.00} & \meanAndVar{2.587}{.36} \\
\rnnLabel{} & \meanAndVar{0.923}{.02} & \meanAndVar{0.000}{.00} & \meanAndVar{0.000}{.00} & \meanAndVar{-24.889}{3.96} & \meanAndVar{0.124}{.01} & \meanAndVar{0.145}{.02} & \meanAndVar{-0.152}{.15} \\
\rnnLabel{}+\supLabel{} & \meanAndVar{0.914}{.01} & \meanAndVar{0.000}{.00} & \meanAndVar{0.000}{.00} & \meanAndVar{-22.570}{3.89} & \meanAndVar{0.114}{.02} & \meanAndVar{0.141}{.01} & \meanAndVar{-0.222}{.19} \\
\rnnLabel{}+\supLabel{}+\readingLabel{} & \meanAndVar{0.932}{.02} & \meanAndVar{0.000}{.00} & \meanAndVar{0.000}{.00} & \meanAndVar{-26.236}{4.04} & \meanAndVar{0.103}{.01} & \meanAndVar{0.112}{.02} & \meanAndVar{-0.083}{.10} \\
\rnnLabel{}+\ndLabel{} & \meanAndVar{0.863}{.08} & \meanAndVar{0.000}{.00} & \meanAndVar{0.000}{.00} & \meanAndVar{-26.566}{6.41} & \meanAndVar{0.107}{.02} & \meanAndVar{0.118}{.01} & \meanAndVar{-0.107}{.15} \\
\rnnLabel{}+\ndLabel{}+\readingLabel{} & \meanAndVar{0.866}{.04} & \meanAndVar{0.000}{.00} & \meanAndVar{0.000}{.00} & \meanAndVar{-17.452}{5.31} & \meanAndVar{0.090}{.01} & \meanAndVar{0.137}{.04} & \meanAndVar{-0.383}{.31} \\
\lstmLabel{} & \meanAndVar{0.224}{.17} & \meanAndVar{0.000}{.00} & \meanAndVar{0.000}{.00} & \meanAndVar{-3.602}{11.43} & \meanAndVar{0.779}{.19} & \meanAndVar{0.066}{.01} & \meanAndVar{2.451}{.42} \\
\lstmLabel{}+\supLabel{} & \meanAndVar{0.111}{.13} & \meanAndVar{0.000}{.00} & \meanAndVarBold{0.000}{.00} & \meanAndVar{8.956}{7.02} & \meanAndVar{0.369}{.12} & \meanAndVar{0.073}{.01} & \meanAndVar{1.552}{.44} \\
\lstmLabel{}+\supLabel{}+\readingLabel{} & \meanAndVar{0.120}{.15} & \meanAndVar{0.000}{.00} & \meanAndVar{0.000}{.00} & \meanAndVar{10.146}{8.82} & \meanAndVar{0.588}{.16} & \meanAndVar{0.065}{.01} & \meanAndVar{2.162}{.23} \\
\lstmLabel{}+\ndLabel{} & \meanAndVar{0.124}{.14} & \meanAndVar{0.000}{.00} & \meanAndVar{0.000}{.00} & \meanAndVarBold{13.303}{13.21} & \meanAndVar{0.640}{.17} & \meanAndVar{0.067}{.01} & \meanAndVar{2.214}{.36} \\
\lstmLabel{}+\ndLabel{}+\readingLabel{} & \meanAndVar{0.045}{.00} & \meanAndVar{0.000}{.00} & \meanAndVar{0.000}{.00} & \meanAndVar{0.419}{6.37} & \meanAndVar{0.500}{.25} & \meanAndVar{0.072}{.02} & \meanAndVar{1.807}{.62} \\
\midrule
mT5\muellerFootnote{}\pretrainedFootnote{} & 1.00 && \text{--} && \text{--} && \text{--} && 1.00\!\!\!* && \text{--} && \text{--} & \\
mBART\muellerFootnote{}\pretrainedFootnote{} & 1.00 && \text{--} && \text{--} && \text{--} && 0.98\!\!\!* && \text{--} && \text{--} & \\
\bottomrule
\end{tabular}

    \caption{\label{tab:passivization-in-german}
        Results on the German passivization task of \citet{mueller-etal-2022-coloring} with certain bugs in the generalization set fixed. Note that the hierarchical and linear partial accuracy scores can sum to more than 1, because the generalization set includes examples where both rules move the same noun type. \muellerFootnoteText{} \pretrainedFootnoteText{} *Since our experiments use a slightly modified generalization set with bugs fixed, these results from prior work are not directly comparable.
    }
\end{table*}

\end{document}